\documentclass[sigconf]{acmart}
\usepackage{hyperref}
   
\usepackage{amsmath}
\usepackage{amsthm}
\usepackage{amssymb}
\usepackage{amsfonts}
\usepackage{graphicx}
\usepackage{natbib}
\usepackage{subfig}
\usepackage{multirow}
\usepackage{amssymb}
\usepackage{mwe}
\usepackage{float}
\usepackage{tabularx,booktabs}
\usepackage{bbold}
\usepackage{mathtools}
\newcolumntype{Y}{>{\centering\arraybackslash}X}

\usepackage{xcolor}
\usepackage{bbm}

\newcommand{\beginsupplement}{%
        \setcounter{table}{0}
        \renewcommand{\thetable}{S\arabic{table}}%
        \setcounter{figure}{0}
        \renewcommand{\thefigure}{S\arabic{figure}}%
     }
     
\newcommand{\E}{\mathbb{E}}
\newcommand{\EE}{\mathbb{E}}
\newcommand{\BR}{\mathbb{R}}
\newcommand{\ud}{\text{d}}
\newcommand{\KL}{\text{KL}}

\newcommand{\ELBO}{\text{ELBO}}

\newcommand{\beq}{\begin{equation}}
\newcommand{\eeq}{\end{equation}}
\newcommand{\beqs}{\begin{eqnarray}}
\newcommand{\eeqs}{\end{eqnarray}}
\newcommand{\barr}{\begin{array}}
\newcommand{\earr}{\end{array}}
\newcommand{\bal}{\begin{align}}
\newcommand{\eal}{\end{align}}

\AtBeginDocument{%
  \providecommand\BibTeX{{%
    \normalfont B\kern-0.5em{\scshape i\kern-0.25em b}\kern-0.8em\TeX}}}

\setcopyright{none}
\renewcommand\acmConference[1]{}
\renewcommand\footnotetextcopyrightpermission[1]{}
\setcopyright{none}
\makeatletter
\renewcommand\@formatdoi[1]{\ignorespaces}

\makeatother

\begin{document}

\title{Variational Learning of Individual Survival Distributions}

\author {Zidi Xiu, Chenyang Tao, Benjamin A. Goldstein, Ricardo Henao}
\affiliation{%
  \institution{Duke University\\
  \{zidi.xiu, chenyang.tao, ben.goldstein, ricardo.henao\}@duke.edu}
%   \city{Durham}
%   \state{North Carolina}
%   \postcode{27705}
}

% % \authornote{Both authors contributed equally to this research.}
% % 
% % \orcid{1234-5678-9012}
% % \author{G.K.M. Tobin}
% % \email{webmaster@marysville-ohio.com}
% \author{Chenyang Tao}
% \email{chenyang.tao@duke.edu}
% \affiliation{%
%   \institution{Duke University}
%   \city{Durham}
%   \state{North Carolina}
%   \postcode{27705}
% }
% \author{Benjamin Goldstein}
% \email{ben.goldstein@duke.edu}
% \affiliation{%
%   \institution{Department of Biostatistics \& Bioinformatics, \\ Duke University}
%   }
% \author{Ricardo Henao}
% \email{ricardo.henao@duke.edu}
% \affiliation{%
%   \institution{ Department of Biostatistics \& Bioinformatics, \\Duke University}
% %   \streetaddress{P.O. Box 1212}
% %   \city{Durham}
% %   \state{North Carolina}
% %   \postcode{27705}
% }

\renewcommand{\shortauthors}{Xiu et al.}


\begin{abstract}
The abundance of modern health data provides many opportunities for the use of machine learning techniques to build better statistical models to improve clinical decision making. Predicting time-to-event distributions, also known as survival analysis, plays a key role in many clinical applications. We introduce a variational time-to-event prediction model, named Variational Survival Inference (VSI), which builds upon recent advances in distribution learning techniques and deep neural networks. VSI addresses the challenges of non-parametric distribution estimation by ($i$) relaxing the restrictive modeling assumptions made in classical models, and ($ii$) efficiently handling the censored observations, {\it i.e.}, events that occur outside the observation window, all within the variational framework. To validate the effectiveness of our approach, an extensive set of experiments on both synthetic and real-world datasets is carried out, showing improved performance relative to competing solutions.

\end{abstract}

%   <concept>
%       <concept_id>10010405.10010444.10010449</concept_id>
%       <concept_desc>Applied computing~Health informatics</concept_desc>
%       <concept_significance>500</concept_significance>
%       </concept>
%   <concept>
%       <concept_id>10010147.10010341.10010342.10010343</concept_id>
%       <concept_desc>Computing methodologies~Modeling methodologies</concept_desc>
%       <concept_significance>300</concept_significance>
%       </concept>
%  </ccs2012>

% \ccsdesc[500]{Applied computing~Health informatics}
% \ccsdesc[300]{Computing methodologies~Modeling methodologies}


\maketitle

\section{INTRODUCTION}
Prediction of event times, also known as {\it survival analysis} in the clinical context, is one of the most extensively studied topics in the statistical literature, largely due to its significance in a wide range of clinical and population health applications.
It provides a fundamental set of tools to statistically analyze the future behavior of a system, or an individual.
In the classical setup, the primary goal of time-to-event modeling is to either characterize the distribution of the occurrence of an event of interest on a population level \citep{kaplan1958nonparametric,kalbfleisch2011statistical}, or more specifically, to estimate a risk score on a subject level \citep{cox1972regression}.
In recent years, there has been a surge of interest in the prediction of individualized event time distributions \citep{yu2011learning}.

A characteristic feature in the study of time-to-event distributions is the presence of censored instances, which refer to an event that is not reported during the follow-up period of a subject.
This can happen, for instance, when a subject drops out during the study (right censoring), including when the study terminates before the event happens (administrative censoring).
Unlike many conventional predictive models, where incomplete observations are usually safely ignored, censored observations contain crucial information that should be adequately considered.
To efficiently leverage the censored observations, together with the complete observations, a classical treatment is to work with the notion of a {\it hazard function}, formally defined as the instantaneous event risk at time $t$, which can be computed by contrasting the event population to the population at risk at a specific time.
Estimates can be derived, for instance by optimizing the {\it partial likelihood} defined by the relative hazards in the case of the Cox Proportional Hazard model (CoxPH) \citep{cox1972regression}.
Alternatively, other work follows the standard {\it Maximal Likelihood Estimation} (MLE) framework, where the individual event distribution is a deformed version of some baseline distribution.
For example, in the {\it Accelerated Failure Time} model (AFT) \citep{kalbfleisch2011statistical}, covariate effects are assumed to rescale the {\it temporal index} of event-time distributions, {\it i.e.}, they either accelerate or delay event progression.
For censored events, their likelihoods are given as the cumulative density after the censoring time \citep{aalen2008survival}.

While vastly popular among practitioners, these models have been criticized for a number of reasons, in particular for the assumptions they make, that consequently render them unfit for many modern applications \citep{wang2019machine}.
For instance, most survival models, including CoxPH and the proportional odds model \citep{murphy1997maximum}, work under the premise of fixed covariate effects, overlooking individual uncertainty.
However, it has been widely recognized that, individual heterogeneity and other sources of variation are common and often time-dependent \citep{aalen1994effects}.
In real-world scenarios, these random factors are typically costly to measure, if not impossible to observe. 
Unfortunately, many models are known to be sensitive to the violation of this fixed effect assumption, raising seriously concerns when deployed in actual practice \citep{hougaard1995frailty}.

Alternatively, machine learning techniques have been leveraged to overcome the limitations of standard statistical survival modeling schemes, especially in terms of model flexibility to address the complexity of data. For example, survival trees employed special node-splitting strategies to stratify the population and derive covariate-based survival curves \citep{bou2011review}, support vector machines \citep{khan2008support} and neural networks \citep{faraggi1995neural} have been used for more expressive predictors and LASSO-type variants \citep{zhang2007adaptive} simultaneously execute variable selection to boost statistical efficiency.
Bayesian statistics has also been explored in the context of model selection \citep{lisboa2003bayesian}, averaging \citep{raftery1995bayesian} and imposing prior beliefs \citep{fard2016bayesian}.
Recent advances in modern machine learning bring extra traction to the concept of data-driven survival models, an important step toward precision medicine.
Prominent examples include direct deep learning extensions of CoxPH \citep{katzman2016deep, li2019deep}, accelerated failure time \citep{chapfuwa2018adversarial} and Bayesian exponential family models \citep{ranganath2016deep}.
Other efforts include the use of Gaussian Process to capture complex interactions between covariates in relation to event times \citep{fernandez2016gaussian} and competing risks \citep{alaa2017deep}.
It has been argued that direct modeling of the event distribution might be beneficial \citep{yu2011learning}, and more recently, adversarial distribution matching has also been considered for survival applications \citep{chapfuwa2018adversarial} with promising results reported.

In this work we present a principled approach to address the challenges of nonparametric modeling of time-to-event distributions in the presence of censored instances.
Our approach, named {\it Variational Survival Inference} (VSI), builds upon recent developments in black-box variational inference \citep{ranganath2014black}.
It directly targets the estimation of individualized event-time distributions, rather than a risk score that correlates with event ordering.
By explicitly accounting for latent variables in its formulation, VSI better accommodates for individual uncertainty.
The proposed VSI is a highly scalable and flexible framework without strong assumptions, featuring easy implementation, stable learning, and importantly, it does not rely on {\it ad-hoc} regularizers.
Our key contributions include:
($i$) a variational formulation of nonparametric time-to-event distribution modeling conditioned on explanatory variables;
($ii$) a cost-effective treatment of censored observations;
($iii$) a thorough discussion on how our modeling choices impact VSI performance, and
($iv$) an empirical validation confirming that the proposed VSI compares favorably to its counterparts on an extensive set of tasks, covering representative synthetic and real-world datasets.

\section{BACKGROUND}
A dataset for survival analysis is typically composed of a collection of triplets $D=\{ Y_i=(t_i, \delta_i,X_i)\}_{i=1}^N$, where $i$ indexes the subjects involved in the study.
For each triplet, $X_i \in \BR^p$ denotes the set of explanatory variables, $t_i$ is the observation time and $\delta_i$ is the event indicator.
To simplify our discussion, we only consider the standard survival setup.
This means $\delta_i$ is binary with $\delta_i=1$ indicating the event of interest happened at $t_i$, otherwise $\delta_i=0$ corresponds to a censoring event, {\it i.e.}, no event occurs until $t_i$ and the subject is unobserved thereafter.
This distinction creates a natural partition of the dataset $D=D_{c} \bigcup D_{e}$, with $D_c= \{Y_i:\delta_i=0\}$ and $D_{e}= \{Y_i:\delta_i=1\}$ representing the censored and event groups, respectively. 
%
%{\bf Notations} For subject $i$, $i=1,\ldots,n$, we observed $Y_i=(t_i, \delta_i,X_i)$ with $t_i=\text{min}(T_i, C_i)$ : the time-to-event or censoring
%, $\delta_i$ indicator of event ($\Delta_i =1$), $X_i$ covariates with subject $i$, with dimension $1\times p$.
%Let $D=\{D_{c}, D_{e}\}$ denotes all data points, with $D_c:= \{Y_i:\Delta_i=0\}$ and $D_{e}:= \{Y_i:\Delta_i=1\}$ for censoring and events group 

% \vspace{-5pt}
\subsection{Statistical survival analysis}
% \vspace{-5pt}
%
In survival analysis, one is interested in characterizing the survival function $S(t)$, defined as the probability that any given subject survives until time $t$.
% proportion of population / the probability of a subject survives until time $t$.
The basic descriptors involved in the discussion of survival analysis are: the cumulative survival density $F(t) = 1-S(t)$, the survival density $f(t) = \partial_t F(t)$, the hazard function $h(t) = \lim_{\Delta t\rightarrow 0} \frac{P(t\leq T < t+\Delta t|T\geq t)}{\Delta t}$ and the cumulative hazard function $H(t) = \int_{0}^t h(s) \ud s$.
The following expressions are fundamental to survival analysis \citep{aalen2008survival}:
%
%\beq
$S(t) = \exp(-H(t))$ and $f(t) = h(t)S(t)$.
% \eeq
%
Further, we use $S(t|x)$, $f(t|x)$, $F(t|x)$, $h(t|x)$, $H(t|x))$ to denote their individualized (subject-level) counterparts given explanatory variables $x$.
All survival models leverage these definitions to derive population-level estimators or subject-level predictive functions, {\it e.g.}, of risk, $S(t|x)$, or event time, $f(t|x)$. 

\subsection{Variational inference}
% \vspace{-5pt}
%
% CAN USE SOME EDITS HERE
For a latent variable model $p_{\theta}(x,z)=p_{\theta}(x|z)p(z)$, we consider $x\in \BR^p$ as an observation, {\it i.e.}, data, and $z\in \BR^m$ as latent variable. 
% The marginal $\log$-likelihood $\log p_{\theta}(x)$ is given by $\log \int p_{\theta}(x,z) \ud z$
The marginal likelihood, given by $p_{\theta}(x) = \int p_{\theta}(x,z) \ud z$, typically does not enjoy a closed form expression.
To avoid direct numerical estimation of $p_{\theta}(x)$, Variational Inference (VI) optimizes a variational bound to the marginal log-likelihood.
The most popular choice is known as the Evidence Lower Bound (ELBO) \citep{wainwright2008graphical}, given by
\beq\label{eq:elbo}
\text{ELBO}(x) 
\triangleq \EE_{Z\sim q_{\phi}(z|x)}
\left[ \log \frac{p_{\theta}(x,Z)}{q_{\phi}(Z|x)} \right] \leq \log p_{\theta}(x), 
\eeq
where $q_\phi(z|x)$ is an approximation to the true (unknown) posterior $p_{\theta}(z|x)$, and the inequality is a direct result of Jensen's inequality.
The variational gap between the ELBO and true $\log$-likelihood is the KL-divergence between posteriors, {\it i.e.}, $\KL(q_{\phi}(z|x)\parallel p_{\theta}(z|x)) = \EE_{q_{\phi}(z|x)}[\log q_{\phi}(z|x) - \log p_{\theta}(z|x)]$, which implies the ELBO tightens as $q_{\phi}(z|x)$ approaches the true posterior $p_{\theta}(z|x)$.
For estimation, we seek parameters $\theta$ and $\phi$ that maximize the ELBO.
At test time, $q_{\phi}(z|x)$ is used for subsequent inference tasks on new data.
% and the commensurately learned parameters $\phi$ are often used in a subsequent inference task with new data.
Given a set of observations $\{ x_i \}_{i=1}^N$ sampled from data distribution $x\sim p_{d}$, maximizing the expected ELBO is also equivalent to minimizing the KL-divergence $\KL(p_d \parallel p_{\theta})$ between the empirical and model distributions.
When $p_{\theta}(x|z)$ and $q_{\phi}(z|x)$ are specified as neural networks, the resulting architecture is more commonly known as the Variational Auto-Encoder (VAE) \citep{kingma2013auto} in the context of computational vision and natural language processing.

% \vspace{-5pt}
\section{VARIATIONAL SURVIVAL INFERENCE}
% \vspace{-5pt}
\label{sec:VSI}
Below we detail the construction of the Variational Survival Inference (VSI) model, which results in predictions of the time-to-event distribution $p_{\theta}(t|x)$ given attribute $x$, with the individual uncertainty accounted in the form of a latent variable $z$ whose distribution is estimated under the VI framework.
Unlike classical survival models, we do not need to specify a parametric form for the baseline distribution, {\it e.g.}, the base hazard $h_0(t)$ in CoxPH \citep{cox1972regression} or the base density $p_0(t)$ in AFT \citep{kalbfleisch2011statistical}.
Instead, we leverage the power of deep neural networks to amortize the learning of the event time and survival distributions, allowing arbitrary (high-order) interactions between the predictors and survival time to be captured.
This overcomes the limitations caused by the restrictive assumptions made in the classical statistical survival analysis frameworks, thus allowing flexible inference of time-to-event distributions.
% A graphical model of our approach is presented in Figure \ref{fig:graphcompare}.
% Here we propose a variational inference based time-to-event distribution learning method (cVAE). The structure is shown in Fig.\ref{fig:graphcompare}.\par

\iffalse
\begin{figure}[t]
\centerline{\includegraphics[scale = 0.3]{plots/graphcompare.pdf}}
\caption{Graphical model for standard VAE (left) and SVI model(right).}
}
\label{fig:graphcompare}    
\end{figure}
\fi

% \vspace{-5pt}
\subsection{Variational bound of observed events}
% \vspace{-5pt}
%
We start the discussion with the simplest scenario, that for which there are no censoring events.
Our goal is to maximize the expected $\log$-likelihood $1/N \sum_i \log p_{\theta}(t_i | X_i)$.
To model the conditional likelihood, we consider a latent variable model of the form $p_{\theta}(t,z|x)$.
The unconditional formulation of the ELBO in \eqref{eq:elbo} can be readily generalized to case conditional on event times as
% , and it can be readily generalized from the unconditional case to derive the following conditional ELBO
%\beqs
%\hspace{-20mm} \ELBO(t|x,p_{\theta},q_{\phi}) = \nonumber\\ 
%\hspace{20mm} \EE_{Z\sim q_{\phi}(z|x,t)}\left[ \log p_{\theta}(t,Z) - \log q_{\phi}(Z|x,t) \right]
%\eeqs
%
\beq\label{eq:elbo_joint}
\ELBO(t|x) =  \EE_{Z\sim q_{\phi}(z|x,t)}\left[ \log \frac{p_{\theta}(t,Z|x)}{q_{\phi}(Z|x,t)} \right],
\eeq
where $q_{\phi}(z|x,t)$ denotes the conditional posterior approximation to the true (unknown) $p_{\theta}(z|x,t)$. 

In particular, we assume a model distribution with the following decomposition
\beq\label{eq:cond}
p_{\theta}(t,z|x) = p_{\theta}(t|z,x) p_{\theta}(z|x) = p_{\theta}(t|z) p_{\theta}(z|x), 
\eeq
which posits that $z$ is a sufficient statistics of $x$ w.r.t. survival time $t$.
Another key assumption we make is that, unlike in the standard variational inference model, we have used a learnable inhomogeneous prior $p_{\theta}(z|x)$ for the latent $z$ to replace the standard fixed homogeneous prior $p(z)$.
Such covariate-dependent prior formulation allows the model to account for individual variation, thus further helping to close the variational gap \citep{tomczak2017vae}.
% {\bf Such practice acknowledges the individual difference in terms of the uncertainties the can be predicted from the covariates, and the use of a more flexible prior further helps to close the variational gap \citep{tomczak2017vae}. }
Replacing \eqref{eq:cond} into the ELBO expression in \eqref{eq:elbo_joint} results in the usual likelihood and KL decomposition pair
\begin{align}
\begin{aligned}
%ELBO(x) = \EE_{Z\sim q_{\phi}}\left[ \log p_{\theta}(Z|x) \right] \\
%\hspace{5mm} - \KL(q_{\beta}(z|x,t) \parallel p_{\theta}(z|x)). 
\ELBO(t|x) & = \EE_{Z\sim q_{\phi}(z|x,t)}\left[ \log p_{\theta}(t|Z) \right] \\[10pt]
& - \KL(q_{\phi}(z|x,t) \parallel p_{\theta}(z|x)),
\end{aligned}
\end{align}
from which we can see that maximizing the ELBO is equivalent to estimate the parameters of a probabilistic time-to-event model $p_{\theta}(t|z)p_{\theta}(z|x)$ with maximum likelihood such that the inhomogeneous prior $p_{\theta}(z|x)$ matches as well as possible a conditional posterior that explicitly accounts for event times, $q_{\phi}(z|x,t)$. 
At test time, only $p_{\theta}(z|x)$ will be used to make predictions provided that $t$ is not available during inference.

More specifically, $p_{\theta}(t|z)$, $p_{\theta}(z|x)$ and $q_{\phi}(z|x,t)$ are defined as neural networks
\begin{align}
    \begin{aligned}
        p_{\theta}(t|z) & = {\rm Softmax}(g(z;\theta)), \\
        p_{\theta}(z|x) & = {\mathcal N}( \mu_p(x;\theta), \Sigma_p(x;\theta)), \\
        q_{\phi}(z|x,t) & = {\mathcal N}( \mu_q(x,t;\phi), \Sigma_q(x,t;\phi) ) ,
        % p_{\theta}(z|x) & = {\cal N}( \mu_p = f_{p,\mu}(x;\theta), \Sigma_p = f_{p,\Sigma}(x;\theta) ) \\
        % q_{\phi}(z|x,t) & = {\cal N}( \mu_q = f_{q,\mu}(x,t;\phi), \Sigma_q = f_{q,\Sigma}(x,t;\phi) ) ,
    \end{aligned}
\end{align}
%
% where $g(z;\theta)$, $f_{p,\mu}(x;\theta)$, $f_{p,\Sigma}(x;\theta)$ and $f_{q,\mu}(x,t;\phi)$, $f_{q,\Sigma}(x,t;\phi)$ 
where $p_{\theta}(t|z)$ is represented on a discretized time line (see below for details), $g(z;\theta)$, $\mu_p(x;\theta)$, $\Sigma_p(x;\theta)$ and $\mu_q(x,t;\phi)$, $\Sigma_q(x,t;\phi)$ are deep neural nets parameterized by model parameters $\theta$ and variational parameters $\phi$, and ${\mathcal N} (\mu, \Sigma)$ denotes the multivariate Gaussian with mean $\mu$ and (diagonal) covariance $\Sigma$.
For standard tabular data, we use Multi Layer Perceptrons (MLPs) to specify these functions. 
% $g(z;\theta)$, $f_{p,\mu}(x;\theta)$ and $f_{p,\Sigma}(x;\theta)$ are Multi Layer perceptrons (MLPs) parameterized by $\theta$, and $f_{q,\mu}(x,t;\phi)$ and $f_{q,\Sigma}(x,t;\phi)$ are MLPs parameterized by $\phi$.

% \vspace{-5pt}
\subsection{Variational bound of censored events}
% \vspace{-5pt}
%
Addressing censoring in the formulation is more challenging as this type of {\em partial observation} is not subsumed in the conventional VI framework.
To address this difficulty, we recall that in likelihood-based survival analysis, the likelihood function for censored observations is given by $\log S_{\theta}(t|x)$, where $S_{\theta}(t|x)$ is the survival function and $t$ is the censoring time.
For censored observations $Y$ with $\delta=0$, we do not have the exact event time $t$.
This means that we only have partial information of the events, in that the event should happen only after the censoring time $t$.
% We could consider using the likelihood for censoring based on the predicted $t$ with the VSI model learned with event observations only.
% \rh{don't know what this means}

% Consider the likelihood for censoring observations.
To derive a tractable objective for censored observations, we first expand $\mathcal{L}_c(x, t) = \log S_{\theta}(t|x)$ based on its definition and an application of Fubini's theorem \citep{resnick2003probability} and Jensen's inequality, {\it i.e.},
\begin{align*}
    \mathcal{L}_c(x, t) & = \text{log }S_{\theta}(t|x)
    = \text{log } \int_{t}^{\infty}p_{\theta}(t|x) dt \\
    & \geq \mathbb{E}_{q_{\phi}(z|t,x)}\left[\text{log }{\frac{p_{\theta}(z|x)}{q_{\phi}(z|t,x)}} + \text{log }S_{\theta}(t|z)\right] \\
    & = \mathbb{E}_{q_{\phi}(z|t,x)}[\text{log }S_{\theta}(t|z)] \\
    & \hspace{15mm} - {\rm KL}(q_{\phi}(z|t,x)||p_{\theta}(z|x))  \\
    &\triangleq\text{ELBO}_c(t|x)
\end{align*}
where the censored log-likelihood bound $\text{ELBO}_c(t|x)$ is only evaluated on $D_c$, {\it i.e.}, the subset of censored observations.
See Supplementary Materials for the full derivation of $\text{ELBO}_c(t|x)$.

\iffalse
\begin{equation}
\begin{split}
\log \mathcal{L}_c(x_i, t_i;\alpha)&=\text{log }S_{\alpha}(t_i|x_i)= \text{log } \int_{t_i}^{\infty}p_{\alpha}(t|x_i) dt\\
%&= \text{log } \int_{t_i}^{\infty}\int_z p_{\alpha}(t,z|x_i) dzdt\\
&= \text{log } \int_{t_i}^{\infty}\int_z \frac{p_{\alpha}(z, t|x_i)}{q_{\beta}(z|t_i,x_i)} q_{\beta}(z|t_i,x_i) dzdt\\
&= \text{log } \int_{t_i}^{\infty}\int_{q_z} \frac{p_{\alpha}(z, t|x_i)}{q_{\beta}(z|t_i,x_i)}  dQ_zdt \\%\text{, by Fubini's theorem}\\
&= \text{log } \int_{q_z} \int_{t_i}^{\infty}\frac{p_{\alpha}(z, t|x_i)}{q_{\beta}(z|t_i,x_i)} dt dQ_z\\
%&= \text{log } \int_{q_z} \int_{t_i}^{\infty}p_{\alpha}(t|z)\frac{p_{\alpha}(z|x_i)}{q_{\beta}(z|t_i,x_i)} dt dQ_z\\
&= \text{log } \int_{q_z} \frac{p_{\alpha}(z|x_i)}{q_{\beta}(z|t_i,x_i)}\int_{t_i}^{\infty}p_{\alpha}(t|z) dt dQ_z\\
&\ge \int_{q_z} \text{log } \left(\frac{p_{\alpha}(z|x_i)}{q_{\beta}(z|t_i,x_i)}\int_{t_i}^{\infty}p_{\alpha}(t|z) dt \right)dQ_z\\
&=  \mathbb{E}_{q_{\beta}(z|t,x)}[\text{log }{\frac{p_{\alpha}(z|x_i)}{q_{\beta}(z|t_i,x_i)}} + \text{log }S_{\alpha}(t_i|z)] \\
 &= -KL(q_{\beta}(z|t_i,x_i)||p_{\alpha}(z|x_i))\\
 &+ \mathbb{E}_{q_z}[\text{log }S_{\alpha}(t_i|z)] \\
&=\text{ELBO}_c
\end{split}
\label{eq:censorVAElb}
\end{equation}
\fi

% \vspace{-5pt}
\subsection{Implementing VSI}
% \vspace{-5pt}
%
In the current instantiation of the model, we discretize time into $M$ bins spanning the time horizon of the (training) data.
This means that (at inference) $t$ is only known up to the time bin it falls into.
% For the the sake of brevity of our exposition, we make the simplifying assumption that the temporal index is pre-partitioned into $M$ bins so each $t$ is only known up to the bin index it falls into.
We note this is not a restrictive assumption as many survival data is only known up to certain temporal accuracy. 
That said, generalization to continuous observations is fairly straightforward.
For datasets that do have a natural discretization, we leave the choice to the user. 
In this study, we partition the temporal index based on the percentiles of observed event time, while also allowing for an artificial $(M+1)$-th bin to account for event times beyond the full observation window, {\it i.e.}, events happening after the end-of-study as observed in the training cohort. 
% In practice, we set $M$ splits based on the empirical quantiles and also allows for a artificial $M+1$-th bin to account for event times beyond the full observation window, {\it i.e.}, events happening after the end-of-study cut in the training cohort. 

Since both $p_{\theta}(z|x)$ and $q_{\phi}(z|x,t)$ are assumed to be Gaussian, 
% Additionally, we choose both prior $p_{\theta}(z|x)$ and approximate posterior $q_{\phi}(z|x,t)$ to be multivariate Gaussian, with means and covariances $(\mu_p, \Sigma_p)$ and $(\mu_q, \Sigma_q)$, respectively.
the following closed-form expression can be used in the computation of the KL terms above
\beq
\begin{array}{c}
    \hspace{-3em}{\rm KL}(q_{\phi}(z|x,t) \parallel p_{\theta}(z|x))) = \frac{1}{2} \left\{ \text{tr}\left( \Sigma_p^{-1} \Sigma_q \right) + \right.\\
    [5pt]
    \hspace{6em} \left.\left( \mu_p - \mu_q \right)^T \Sigma_p^{-1} \left( \mu_p - \mu_q \right) - m + \log \frac{\det(\Sigma_p)}{ \det(\Sigma_q)}\right\}.
\end{array}
\eeq
Following \citet{ranganath2014black}, we use diagonal covariance matrices and apply the reparameterization trick to facilitate stable differatiable learning. 

%{\bf Get smoothed time-to-event distribution}
In order to compute the term $S_{\theta}(t|x)$, we use discretized time scheme as previously described, and sum up all predicted probabilities subsequent to bin $t$. 
% For integration purpose, discretizing time is not the only way. 
Note that this can be readily generalized to continuous time models.
So long as the cumulative distribution of $p_{\theta}(t|z)$ enjoys a closed form expression, a numerical integration scheme is not necessary to implement VSI. 

% detailed description for implementation can be found in the Supplementary

% \vspace{-5pt}
\subsection{Importance-Weighted estimator for likelihood evaluation}
% \vspace{-5pt}
%
For evaluation purposes, we need to be able to compute the model's log-likelihood for an observation $Y = (x_i,t_i,\delta_i)$, {\it i.e.},
\beq
\mathcal{L}_{\rm VSI}(x_i,t_i;\theta)  = \delta_i\log p_\theta(t_i|x_i) \\
 + (1-\delta_i)\log S_\theta(t_i|x_i). %\\
\eeq
%&= \delta_i \log \prod_{b=1}^B \pi_{(i)b}^{t_{(i)b}} + (1-\delta_i)\log (1-\sum_{b=1}^{\text{argwhich}_k(t_{(i)k}=1)}\pi_{(i)b})
%\end{split}
In this study, we use the importance-weighted (IW) estimator \citep{burda2015importance}, which provides a tighter bound to the $\log$-likelihood. While more sophisticated alternatives might provide sharper estimates \citep{neal2001annealed}, we deem IW estimator sufficient for the scope of this study. Additionally, while the tighter bound can be repurposed for training, it does not necessarily result in improved performance \citep{rainforth2018tighter}, which we find to be the case in this study. 
%
% \begin{align}
%\begin{split}

To obtain a more accurate value of the likelihood, we use the approximate posterior as our proposal, and use the following finite sample estimate
% Note that to obtain a more accurate value of the log-likelihood given that we are estimating a lower bound, we need to consider importance sampling, then calculate the integral with Monte Carlo (MC) sampling as
%
\begin{equation*}
\begin{split}
\hat{p}_\theta(t_i|x_i) &= \int \frac{p_{\theta}(t_i|z)p_{\theta}(z|x)}{q_{\phi}(z|t_i,x)}q_{\phi}(z|t_i,x) dz \\
% p(t_i|x_i)&=\int \frac{p_{\alpha}(t_i,z|x)}{q_{\beta}(z|t_i,x)}q_{\beta}(z|t_i,x) dz\\
% &=\int \frac{p_{\alpha}(t_i|z)p_{\alpha}(z|x)}{q_{\beta}(z|t_i,x)} dQ_{\beta}(z)\\
& \approx \frac{1}{L}\sum_{l=1}^L \frac{p_{\theta}(t_i|z_l)p_{\theta}(z_l|x)}{q_{\phi}(z_l|t_i,x)},
\end{split}
\end{equation*}
where $L$ is the number of samples.
The log-likelihood for the corresponding conditional survival function is
\begin{equation*}
\begin{split}
\hat{S}_\theta(t_i|x_i) &= \int_{t>t_i}\int \frac{p_{\theta}(t_i|z)p_{\theta}(z|x)}{q_{\phi}(z|t_i,x)}q_{\phi}(z|t_i,x) dz dt\\
& \approx \frac{1}{L}\sum_{l=1}^L \frac{ \int_{t>t_i} p_{\theta}(t,z_l|x)dt}{q_{\phi}(z_l|t_i,x)}
\end{split}
\end{equation*}
Note that by nature of Jensen's inequality, the resultant estimand will be an under-estimation of the true $\log$-likelihood. As $L$ goes to infinity, the approximated lower bound will converge to the true $\log$-likelihood.
%
% \vspace{-5pt}
\subsection{Making Predictions}
\label{sec:wtaverage}
% \vspace{-5pt}
%
{\bf Predictive time-to-event distribution}
During inference, given a new data point with $x_*$, according to the generative  model $p_{\theta}(t|x_*) = \int p_{\theta}(t,z|x_*)dz = \int p_{\theta}(t|z)p_{\theta}(z|x_*)dz$, where the integration is conducted numerically by Monte Carlo sampling.

{\bf Point estimation of time-to-event}
To better exploit the learned approximated posterior $q_{\phi}(z|x,t)$, we generalize the importance sampling idea and provide a weighted average as time-to-event summary, rather than for instance using a summary statistic such as median or mean.
% Based on previous part, $\hat{t}_i \sim p_{\alpha}(t|x^{\text{new}})$.
Specifically, consider multiple samples of $t_*^{(l)} \sim p_{\theta}(t|x_*)$, then calculate a weighted average as %\cytao{notation overload, plz fix.}
\beq\label{eq:wt}
\begin{array}{c}
t_* = \frac{\sum_{l=1}^L w_*^{(l)} t_*^{(l)}}{\sum_{l=1}^L w_*^{(l)}},\, w_*^{(l)} = \frac{p_{\theta}(z_l|x_*)}{q_{\phi}(z_l|t_*^{(l)},x_*)}, \\
[10pt]
t_*^{(l)} \sim p_{\theta}(t|x_*), \,\,\, z_l \sim q_{\phi}(z|t_*^{(l)},x_*). 
\end{array}
\eeq
% \begin{equation}\label{eq:wt}
% \begin{split}
% t_* &= \frac{1}{\sum_{l=1}^L w_*^{(l)}}\sum_{l=1}^L w_*^{(l)} t_*^{(l)}\\
% w_*^{(l)} & = \frac{p_{\theta}(z_l|x_*)}{q_{\phi}(z_l|t_*^{(l)},x_*)} \\
% t_*^{(l)} &\sim p_{\theta}(t|x_*) \\
% z_l &\sim q_{\phi}(z|t_*^{(l)},x_*)
% \end{split}
% \end{equation}
%
% \begin{equation}\label{eq:wt}
% \begin{split}
% t_* &= \frac{1}{\sum_{l=1}^L w_*^{(l)}}\sum_{l=1}^L w_*^{(l)} t_*^{(l)}\\
% w_*^{(l)} & = \frac{p_{\theta}(z_l|x_*)}{q_{\phi}(z_l|t_*^{(l)},x_*)} \\
% t_*^{(l)} &\sim p_{\theta}(t|x_*) \\
% z_l &\sim q_{\phi}(z|t_*^{(l)},x_*)
% \end{split}
% \end{equation}
%
In the Supplementary Materials we show that \eqref{eq:wt} gives better model performance for point-estimate-based evaluation metrics, Concordance Index in particular, compared to other popular summary statistic such as the median of $t_* \sim p_{\theta}(t|x_*)$ with $L$ empirical samples. 
% C-Index was calculated based on this point estimation, however C-Index based on median estimates provided in the Supplementary Materials.

% \vspace{-5pt}
\section{DISSECTING VSI}
% \vspace{-5pt}
\label{sec:baselines}
In the experiments, we show the effectiveness of the proposed VSI model in recovering underlying time-to-event distributions.
To provide additional insight into the differentiating components of the VSI model, we consider two baseline models that partially adopt a VSI design, as detailed below. 
% there are not many statistical or deep learning methods that provide subject-level prediction of the time-to-event distribution directly, thus we designed two baseline deep learning models based on the survival likelihood in \eqref{eq:survlikeli} to be used in the experiments as baseline.

\vspace{3pt}
{\bf VSI without a $q_\phi$ arm (VSI-NoQ)}
In VSI, we use the variational lower bound to maximize the likelihood in survival studies by implicitly forcing the unknown intractable model posterior $p_\theta(z|x)$ to be close to the tractable posterior approximation $q_{\phi}(z|x, t)$.
Via the KL divergence minimization, such matching allows the model to better account for interactions between covariates $x$ and event times $t$ captured by $q_{\phi}(z|x,t)$ to better inform the construction of the latent representation $z$ via isolating out the individual uncertainty encoded by $p_{\theta}(z|x)$. 
% , which requires a tractable posterior $q_{\phi}(z|x, t)$ as a substitution of the unknown intractable posterior $p(z|x,t)$.
% Also, the behavior of the tractable posterior $q_{\beta}$ can be regularized with the KL Divergence term in the ELBO for both events in \eqref{eq:VAElb} and censoring in \eqref{eq:censorVAElb}.
If we exclude the interaction term $(x,t)$ in $q_{\phi}$ and only make the prediction with $x$, {\it i.e.}, with the approximate posterior given by $q_{\phi}(z|x)$, through the same stochastic latent representation $z$, then naturally the optimal solution is to equate $q_{\phi}(z|x)$ with the prior $p_{\theta}(z|x)$ \footnote{Based on a KL-vanishing argument.}. This basically eliminates $q_{\phi}$ from our formulation, and therefore we call this variant VSI-NoQ. 

More specifically, without a $q_{\phi}$ arm the model described in Section \ref{sec:VSI} essentially becomes a feed-forward model with a special stochastic hidden layer $z$. 
In this case, the model likelihood is given by $p_{\theta}(t|x) = \int p_{\theta}(t,z|x)dz = \int p_{\theta}(t|z)p_{\theta}(z|x)dz$, where $p_{\theta}(t|z)$ and $p_{\theta}(z|x)$ are defined as in \eqref{eq:cond}.
Note that the only difference with VSI is the lack of the KL divergence term to match $p_{\theta}(z|x)$ to $q_{\phi}(z|x, t)$.
This baseline model (VSI-NoQ) is considered to dissect the impact of excluding complex interaction between covariates and event time when constructing the individualized priors. 
% such matching in performance while still keeping the stochastic latent representation $z$ in the model.

% {\bf VSI without a $q_\phi(\cdot)$ arm}
% In VSI, we use the variational lower bound to approximate the likelihood in survival studies by requiring the unknown intractable inhomogeneous posterior $p_\theta(z|x)$ to match the tractable approximation $q_{\phi}(z|x, t)$.
% Such matching allows the model to better account for interactions between covariates and event times when constructing the latent representation $z$ via KL divergence minimization.
% % , which requires a tractable posterior $q_{\phi}(z|x, t)$ as a substitution of the unknown intractable posterior $p(z|x,t)$.
% % Also, the behavior of the tractable posterior $q_{\beta}$ can be regularized with the KL Divergence term in the ELBO for both events in \eqref{eq:VAElb} and censoring in \eqref{eq:censorVAElb}.
% Without $q_{\phi}(\cdot)$, the model in Section \ref{sec:VSI} becomes a feed-forward model with a stochastic hidden layer $z$. 
% In this case, $p_{\theta}(t|x) = \int p_{\theta}(t,z|x)dz = \int p_{\theta}(t|z)p_{\theta}(z|x)dz$, where $p_{\theta}(t|z)$ and $p_{\theta}(z|x)$ are defined as in \eqref{eq:}.
% Note that the only difference with VSI is the lack of the KL divergence term to match $p_{\theta}(z|x)$ to $q_{\phi}(z|x, t)$.
% This baseline model is considered to evaluate the impact of such matching in performance while still keeping the stochastic latent representation $z$ in the model.

\vspace{3pt}
{\bf Deterministic feed-forward model (MLP)} To understand the importance of the stochastic latent representations $z$, we consider a straightforward baseline which directly predicts the event time distribution based on the input $x$, {\it i.e.}, $p_{\theta}(\cdot|x) = \rm{Softmax}(g_{\theta}(x))$, which is essentially a standard multinomial regression with censored observation. In our study, we use the MLP to implement $g_{\theta}(x)$. And as such, hereafter we will refer to this model as MLP. Additionally, we also considered standard randomization schemes, such as dropout \citep{srivastava2014dropout}, in the construction of a stochastic neural net, which promises to improve performance. Such strategy also incorporates randomness, however differs principally from the modeled uncertainty exploited by our VSI scheme. In our experiment section, we report the best results from MLP with or without dropout.

% {\bf Deterministic feed-forward model}
% Consider a standard MLP specification which deterministically outputs event time distributions conditioned on the inputs $x$ as $p_\theta(t|z)$ for $z=f(x;\theta)$.
% In this setting the model lacks of stochastic latent variables thus optimization is carried out by MLE, {\it i.e.}, directly maximizing \eqref{}.
% , which could also calculate $S(t|x)$ for censoring likelihood.
% This model is another generative method.
% The learning objective is also $\log \mathcal{L}(x_i,t_i;\theta) = \Delta_i \log p(t_i|x_i) + (1-\Delta_i) \log S(t_i|x_i)$, without pre-assumed latent $z$.

These baseline approaches use feed-forward deep learning networks to learn $p_\theta(t|x)$ without incurring the notation of variational inference.
In the experiments we will show that the variational inference is crucial to the accurate learning of time-to-event distributions, resulting in better performance relative to these baselines, especially when the proportion of censoring events is high.
% actually helps learning time-to-event distribution, especially with censoring data.

% \vspace{-5pt}
\section{RELATED WORK}
% \vspace{-5pt}
%

\vspace{3pt}
{\bf Machine learning and survival analysis} Early attempts of combining machine learning techniques with statistical survival analysis, such as the Faraggi-Simon network (FS-network) \citep{faraggi1995neural}, often failed to demonstrate a clear advantage over classical baselines \citep{schwarzer2000misuses}. Recent progresses in machine learning allow researchers to overcome the difficulties suffered by prior studies. For example, \citet{katzman2018deepsurv} showed that weight decay, batch normalization and dropout significantly improved the performance of FS-network. \citet{li2019deep} analyzed survival curves based on clinical images using deep convolution neural net (CNN). In addition to deep nets, \citet{fernandez2016gaussian} showed that Gaussian Process can be used to effectively capture the non-linear variations in CoxPH models,  and \citet{alaa2017deep} further proposed a variant that handles competing risks. Similar to these works, our VSI also draws power from recent advances in machine learning to define a flexible learner. 

\vspace{3pt}
{\bf Bayesian survival analysis} Bayesian treatment of survival models has a long history. \citet{raftery1996accounting} first considered modeling uncertainties for survival data, \citet{zupan1999machine} reported probabilistic analysis under Bayesian setup. More recently, \citet{fard2016bayesian} exploited the Bayesian framework to extrapolate priors, and \citet{zhang2018nonparametric} described a Bayesian treatment of competing risks. Closest to VSI is the work of {\it deep exponential family} model (DEF) survival model \citep{ranganath2016deep}, where the authors introduced a Bayesian latent variable model to model both predictors $x$ and survival time $t$. Unlike our VSI, DEF still imposes strong parametric assumptions on the survival distribution, and it's not clear how the censored observations are handled in DEF's actual implementation. Another key difference between DEF and VSI is the factorization of joint likelihood. As the VSI encoder will only seek to capture the latent components that are predictive of the survival time distribution, while DEF encoder also needs to summarize information required to reconstruct covariates $x$. We argue that our VSI factorization of joint probability is more sensible for survival time modeling, because modeling $x$ not only adds model complexity but also introduces nuisance to the prediction of survival time $t$. For datasets with large covariates dimensions and noisy observations, the DEF features can be dominated by the ones predictive of $x$ rather $t$, compromising the main goal of modeling the survival distribution.

\vspace{3pt}
{\bf Individual uncertainties and randomization} The seminal work of \citet{aalen1994effects} first identified importance of accounting for the individual uncertainties, the main culprit for the failure of classical survival models, which can be remedied by explicitly modeling the random effects \citep{hougaard1995frailty}. Alternatively, \citet{ishwaran2008random} presented {\it Random Survival Tree} (RST) to predict cumulative hazards using a tree ensemble, demonstrating the effectiveness of a randomization scheme for statistical survival models. Our approach differs from the above schemes by systematically account for individual uncertainty using the randomness of latent variables.

\vspace{3pt}
{\bf Direct modeling of survival distribution} The pioneering work of \citet{yu2011learning} advocated the prediction of individual survival distributions, which is learned using a generalized logistic regression scheme. This idea is further generalized in the works of \citet{luck2017deep} and \citet{fotso2018deep}. Recently, \citet{chapfuwa2018adversarial} explored the use of deep {\it Generative Adversarial Network} (GAN) to capture the individual survival distribution, which is closest to our goal. Compared the proposed VSI, the adversarial learning of survival distribution is largely unstable, and its success crucially relies on the use of {\it ad-hoc} regularizers.

% To model the time-to-event distribution directly, 

% \citet{yu2011learning} considered survival regression at each discretized time based on generalized logistic regression and \citet{fotso2018deep} substituted the above logistic regression with deep learning networks.
% In \citet{ranganath2016deep}, they modeled the time-to-event with a hierarchical generative approach engaging deep exponential families.
% Both covariates and survival time were specified conditional on a latent process.
% \cite{luck2017deep} focused on subject-specific survival distribution by jointly learning time-to-event and the corresponding rank in the cox partial log-likelihood framework.
% With layer sharing network structure, \citet{lee2018deephit} jointly learned competing risks between different events.
% \citet{chapfuwa2018adversarial} focused on modeling time-to-event directly using Generative Adversarial Network(GAN) as a point estimation problem.
% To get unbiased loss for censoring, \citet{steingrimsson2019deep} used inverse weighting technique to transform the loss of full data deep learning algorithm a censoring unbiased loss functions.

% \vspace{-5pt}
\section{EXPERIMENTS}
% \vspace{-5pt}

%
To validate the effectiveness of the proposed VSI, we benchmarked its performance against the following representative examples from both statistical and machine learning survival analysis schemes: AFT-Weibull, CoxPH, LASSO-based CoxNet \citep{simon2011regularization}, Random Survival Forest (RSF) \citep{ishwaran2008random} and deep learning based DeepSurv \citep{katzman2018deepsurv}.
To fully appreciate the gains from using a variational setup, we further compared the results with the baselines discussed in Section \ref{sec:baselines}, namely, the feed-forward model (MLP) and VSI model without the backward encoding arm $q_{\phi}(z|t,x)$ (VSI-NoQ).

%% Training Details Added
For data preparation, we randomly partition data into three non-overlapping sets for training (60\%), validation (20\%) and evaluation (20\%) purposes respectively. All models are trained on the training set, and we tune the model hyper-parameters wrt the out-of-sample performance on the validation set. The results reported in the paper are based on the evaluation set using best-performing hyper-parameters determined by the validation set. We apply ADAM optimizer with learning rate of $5\times 10^{-4}$ during training, with mini-batches of size $100$. The early stopping criteria of no improvement on the validation datasets is enforced. 

To ensure fair comparisons, all deep-learning based solutions are matched for the number parameters and similar model architectures \& similar hyper-parameter settings. 
TensorFlow code to replicate our experiments can be found at \url{https://github.com/ZidiXiu/VSI/}.
The details of the VSI model setups are related to the Supplementary Materials (SM).

% \vspace{-5pt}
\subsection{Evaluation Metrics}
% \vspace{-5pt}

To objectively evaluate these competing survival models, we report a comprehensive set of distribution-based and point-estimate based scores to assess model performance, as detailed below. 

% is one of the most popular performance metric used in survival studies
\vspace{3pt}
{\bf Concordance Index} (C-Index) is commonly used to evaluate the consistency between the model predicted risk scores and observed event rankings \citep{harrell1982evaluating}. Formally, it is defined as 
\begin{equation*}
\text{C-Index} = \frac{1}{|\mathcal{E}|} \sum_{(i,j)\in\mathcal{E}} \mathbb{1}_{f(x_i)>f(x_j)}    
\end{equation*}
,
% \beq
% \text{C-Index} = \frac{1}{|\mathcal{E}|} \sum_{(i,j)\in\mathcal{E}} \mathbb{1}_{f(x_i)>f(x_j)},
% \eeq
where $\mathcal{E}=\{ t_i < t_j | \delta_i=1 \}$ is the set of all valid ordered pairs (event $i$ before event $j$) and $f(x)$ is a scalar prediction made by the model. Higher is better. 

% {\bf Concordance Index} With C-Index defined in \citet{harrell1982evaluating}, which is an assessment of the discrimination ability of survival analysis models \citep{pencina2004overall}.
% A concordant is a subject risk predicted by the model as higher than another subject who has an event at an later time point.
% The concordant pairs only considering pair ordering among subjects, without time horizon.
\vspace{3pt}
{\bf Time-dependent Concordance Index} is a distribution generalization of the scalar risk score based C-Index \citep{antolini2005time}, which is computed from the predicted survival distribution. Formally it is given by %$\mathcal{C}^{\text{td}} = P(\hat{F}(t_i|x_i)>\hat{F}(t_j|x_j)|t_i<t_j)$ 
\begin{equation*}
    \mathcal{C}^{\text{td}} = P(\hat{F}(t_i|x_i)>\hat{F}(t_i|x_j)|t_i<t_j).
\end{equation*}
, where $\hat{F}$ denotes the model predicted cumulative survival function.
We report the results using the following empirical estimator 
\begin{equation*}
\hat{\mathcal{C}}^{\text{td}} = \frac{1}{|\mathcal{E}|}\sum_{(i,j)\in \mathcal{E}}\mathbb{1}_{\hat{F}(t_i|x_i)>\hat{F}(t_j|x_j)}
\end{equation*}
% \beq
% \hat{\mathcal{C}}^{\text{td}} = \frac{1}{|\mathcal{E}|}\sum_{(i,j)\in \mathcal{E}}\mathbb{1}_{\hat{F}(t_i|x_i)>\hat{F}(t_j|x_j)}
% \eeq

% {\bf Time-dependent Concordance Index}
% According to \cite{antolini2005time} and \cite{wolbers2014concordance}, regarding the original definition of Concordance Index \cite{harrell1982evaluating}, can reflex the concordance based on estimated survival function.
% %
% \begin{equation}
% \mathcal{C}^{\text{td}} = P(\hat{F}(t_i|x_i)>\hat{F}(t_i|x_j)|t_i<t_j)
% \end{equation}
% %
% $\hat{F}(\cdot|x_i)$ is the estimated cumulative distribution for time-to-event with covariates $x_i$. To calculate $\mathcal{C}^{\text{td}}$, consider randomly sampling pairs $(Y_i,Y_j)$ such that $\delta_i=1$, and $Y_j$ does not have event or have been censored before $t_j$. Denote set of those pairs as $V$. Then we can have the empirical $\mathcal{C}^{\text{td}}$ as:
% %
% \begin{equation*}
% \begin{split}
% \mathcal{C}^{\text{td}} &= P(\hat{F}(t_i|x_i)>\hat{F}(t_i|x_j)|t_i<t_j)\\
% &= \E(\mathrm{I}[\hat{F}(t_i|x_i)>\hat{F}(t_i|x_j)|t_i<t_j])\\
% &\approx \frac{\sum_{(i,j)\in V}\mathrm{I}[\hat{F}(t_i|x_i)>\hat{F}(t_i|x_j)]}{|V|}
% \end{split}
% \end{equation*}
% %
% where $|V|$ is the cardinality of $V$.
\vspace{3pt}
{\bf Kolmogorov-Smirnov (KS) distance} For synthetic datasets, we also report the KS distance \citep{massey1951kolmogorov} between the predicted distribution and the ground truth. KS computes the maximal discrepancy between two cumulative densities, {\it i.e.},
\begin{equation*}\text{KS} = \text{sup}_t |F_1(t)-F_2(t)|,
\end{equation*}
% \[
% \text{KS} = \text{sup}_t |F_1(t)-F_2(t)|,
% \]
and a lower KS indicates better match of two distributions. 

\vspace{3pt}
{\bf Test log-likelihood} We also report the average $\log$-likelihood on the held-out test set. A higher score indicates the model is better aligned with the ground-truth distribution in the sense of KL-divergence. Additionally, we also evaluate the spread of empirical likelihood wrt the models. In the case of an expected $\log$-likelihood tie, models with the more concentrated $\log$-likelihoods are considered better under the maximal entropy principle \citep{chen2018variational} ({\it i.e.}, as observed instances received more uniform/similar likelihoods, better generalization of the model is implied).
% {\bf Kolmogorov-Smirnov (KS) distance}
% We could also compare KS statistics \citep{massey1951kolmogorov} for the learned time-to-event distribution, in the simulation cases where we know the ground truth.
% KS computes the statistics $D_n$:
% \[
% D_n = \text{sup}_t |F^{i}_n(t)-F^{ii}_n(t)|,
% \]
% where $D_n$ quantifies the maximum discrepancy of the two CDFs, $F^{i}_n(t)$ and $F^{ii}_n(t)$ in $L_{\infty}$ norm.
% We can also evaluate our method with a summary statistics of the predicted distribution to the true $t$ directly. However in real world, the true CDF is usually unknown, so we only compared this statistic in simulation studies.

\vspace{3pt}
{\bf Coverage Rate}
To quantify the proportion of observed time covered in the predicted personalized time-to-event distributions, we calculated the coverage rate for different percentile ranges. For subjects with event observations, the coverage rate is defined as the proportion of observations fall in the percentile ranges $[l, u]$ of the predicted distributions, where $l,u$ respectively denotes lower and upper quantile of percentile ranges, {\it i.e.},
\begin{equation*}
    \text{Cover Rate}_{\text{events}}(l,u)=\frac{1}{n_e}\sum_{y_i \in \mathcal{D}_{e}}\mathrm{I}(l<t_i<u)
\end{equation*}
In our experiments, we report coverage rates of events at percentile range $[l,u] \in \{ [0.05, 0.95]$, $[0.1, 0.9]$, $[0.15, 0.85]$, $[0.2, 0.8]$, $[0.25, 0.75]$, $[0.3, 0.7]$, $[0.35, 0.65]$, $[0.4, 0.6]$, $[0.45, 0.55]\}$ of the predicted personalized distributions. 
% , $\ldots$, $[0.45, 0.55]\}$.
For censoring, we calculate the proportion of the censoring time happened before the percentiles of predicted range, since the true time-to-event for censoring is happened after censoring time, 
\begin{equation*}
\text{Cover Rate}_{\text{censor}}(l)=\frac{1}{n_c}\sum_{y_i \in \mathcal{D}_{c}}\mathrm{I}(t_i\le l)    
\end{equation*}
We evaluated the coverage rate for censoring at $l \in\{ 0.1, 0.2, \cdots, 0.9 \}$ percentiles.
% $[0.1, 1.0]$, $[0.2, 1.0]$, $\ldots$, $[0.9, 1.0]$. 

For all coverage rates, a higher score implies better performance. Coverage rates for events and censoring should be considered together to evaluate model performance.

% \vspace{-5pt}
\subsection{Synthetic datasets}
% \vspace{-5pt}
Following \citet{bender2005generating} we simulate a realistic survival data based on the German Uranium Miners Cohort Study in accordance with the Cox-Gompertz model $$T = \frac{1}{\alpha}\text{log}\left[1-\frac{\alpha\text{log}(U)}{\lambda \text{exp}(\beta_{\text{age}}\times \text{AGE} + \beta_{\text{radon}}\times \text{RADON})}\right]$$,
% \begin{equation*}
% h(t|x) = \lambda \text{exp}(\alpha t) \text{exp}(\beta_{\text{age}}\times \text{AGE} + \beta_{\text{radon}}\times \text{RADON}).
% \end{equation*}
% \begin{equation}\label{eq:pt_sim}
% T = \frac{1}{\alpha}\text{log}\left[1-\frac{\alpha\text{log}(U)}{\lambda \text{exp}(\beta_{\text{age}}\times \text{AGE} + \beta_{\text{radon}}\times \text{RADON})}\right],
% \end{equation}
with $U \sim \text{Unif}[0,1]$. 
This model simulates the cancer mortality associated with radon exposure and age. Model parameter are derived from real data: $\alpha=0.2138$, $\lambda=7\times 10^{-8}$, $\beta_{\text{age}} =0.15$ and $\beta_{\text{radon}}=0.001$. Covariates are generated according to 
% ${\rm AGE}\sim {\mathcal N}(24.3, 8.4^2)$ and ${\rm RADON}\sim{\mathcal N}(266.8, 507.8^2)$,
\begin{equation*}
{\rm AGE}\sim {\mathcal N}(24.3, (8.4)^2), \, {\rm RADON}\sim{\mathcal N}(266.8, (507.8)^2), 
\end{equation*}
where $\mathcal{N}(\mu,\sigma^2)$ denotes a normal distribution with mean $\mu$ and variance $\sigma^2$. We simulate uniform censoring within a fixed time horizon $c$, {\it i.e.}, we let $C_i \sim \text{UNIF}(0,c)$, then $\delta_i=\mathbb{1}_(T_i < C_i)$ and $T_i=C_i$ if $C_i < T_i$. By setting different upper bounds  $c$ for censoring, we achieve different observed event rates, $100\% (c=\infty)$, $50\%  (c=100)$ and $30\%  (c=70)$.
For each simulation we randomly draw $N=50k$ iid samples.

\begin{figure}[t!]
\centering
\begin{tabular}{c@{}c}
\subfloat[$\delta=1$ (observed)]{\includegraphics[width=0.5\linewidth]{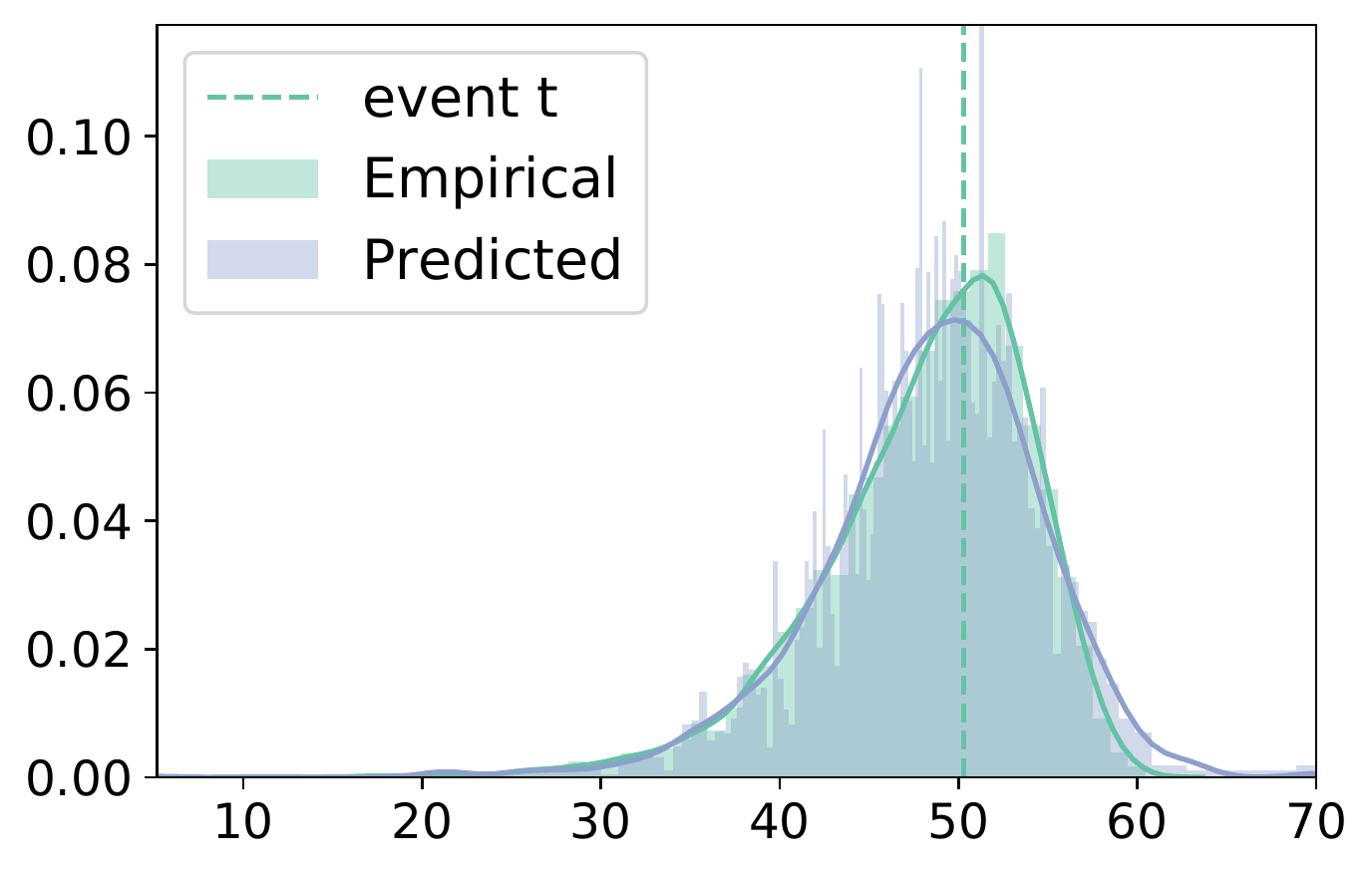}} & \subfloat[$\delta=0$ (censored)]{\includegraphics[width=0.5\linewidth]{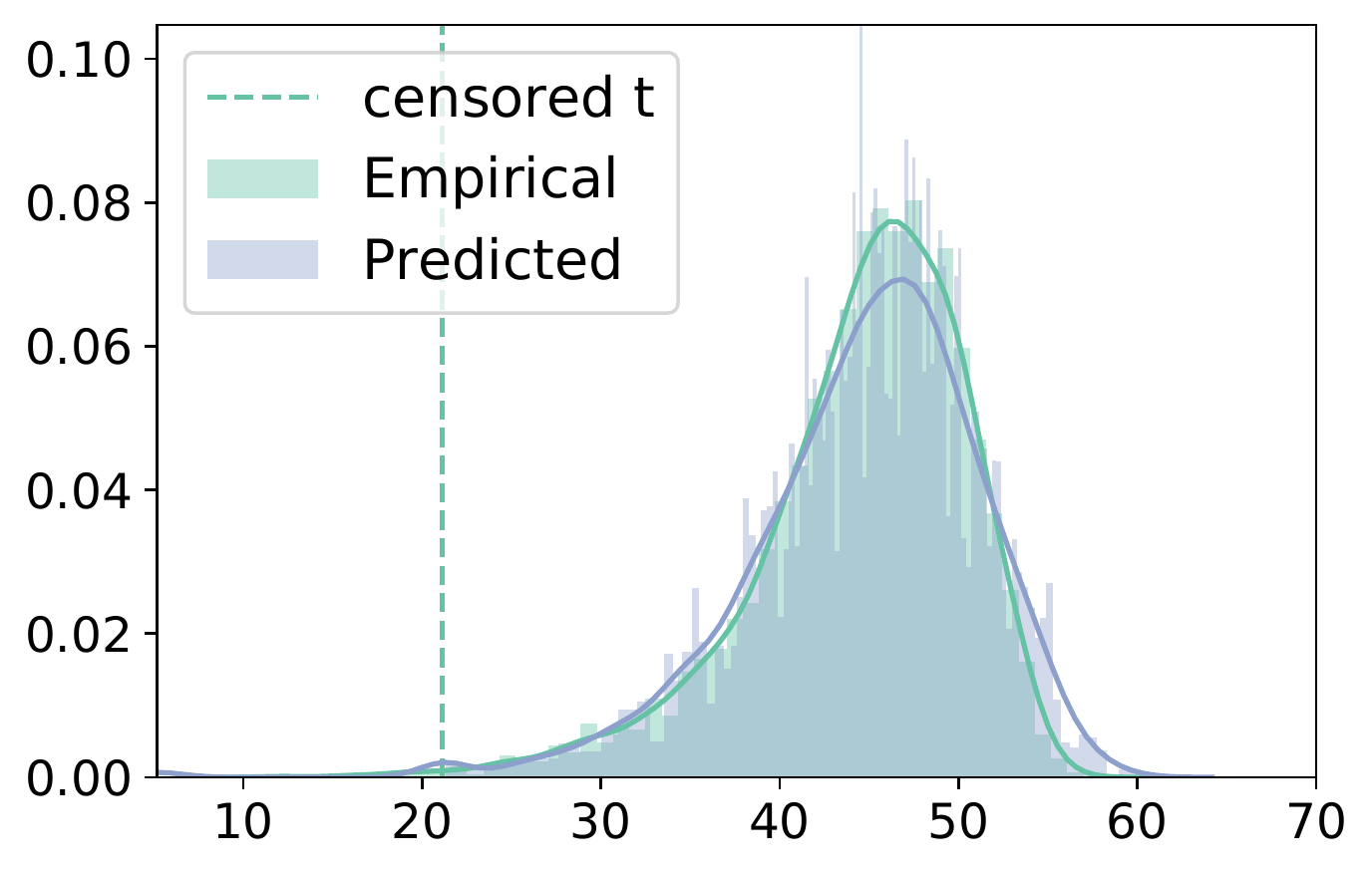}}\\
\end{tabular}
\caption{Two simulated time-to-event distributions with 30\% event rate showing that VSI successfully predicts the underlying distributions from covariates. (left: events, right:censoring)}
% \vspace{-1em}
\label{fig:simusubj}
\end{figure}

\vspace{3pt}
{\bf Prediction of subject-level distribution} In practice, for each subject we only observe one $t$ from its underlying distribution.
Our goal is to accurately predict the underlying distribution from the covariates $x$ alone (since $t$ and $\delta$ are not observed at test time), by learning from the observed instances.
Figure~\ref{fig:simusubj} compares our VSI prediction with the ground-truth for two random subjects, which accurately recovers of individual survival distribution for both observed (Figure~\ref{fig:simusubj}(a)) and censored cases (Figure~\ref{fig:simusubj}(b)).
%
% For each subject we only observe one $t$, and it is a sample from an underlying distribution, our aim is to recover the underlying distribution, as shown in Figure \ref{fig:simusubj} (left). For censoring subjects, which is assumed that censoring time is independent of event time, we can also recover the underlying distribution (Figure \ref{fig:simusubj} (right)). \par

\begin{table}[ht]
\centering
\caption{KS statistic for simulation study.}
\begin{tabular}{lrrr}
\toprule
Event Rate  & 100\%   & 50\%     & 30\%    \\ \midrule
CoxPH (Oracle) & 0.027 &0.032  & 0.027\\
[5pt]
AFT-Weibull & 0.057 & 0.058  & 0.068\\
MLP         & 0.047 & 0.063  & 0.064 \\
[5pt]
VSI-NoQ    & 0.049 & 0.068 & 0.066 \\
VSI        & \textbf{0.044} & \textbf{0.052}  & \textbf{0.059} \\
% \hline
\bottomrule
\end{tabular}
\label{Tab:simulationKS}
\vspace{-1em}
\end{table}
To systematically evaluate the consistency between the predicted and the true distributions, we compare average KS distance from models trained with various event rates in Table~\ref{Tab:simulationKS}.
Since the underlying generative process is based on CoxPH model, we consider the results from CoxPH as the oracle reference, since there is no model mis-specification.
% for each case as shown in Tab.\ref{Tab:simulationKS}. Since this is a model totally based on CoxPH assumptions, CoxPH model's KS statistics are the best we could have which serve as the reference.
At 100\% event rate ({\it i.e.}, complete observation), apart from the oracle CoxPH, all models perform similarly. The VSI variants give slightly better results compared with MLP and AFT-Weibull.
As the proportion of observed events decreases, VSI remains the best performing model, closely followed by the parametric AFT-Weibull.
Note that neither MLP nor VSI-NoQ matches the performance of VSI, which suggests that the full VSI design better accommodates censoring observations.
% the mean KS statistics of the test dataset. With decreasing event rate, VSI remains a robust prediction of the time-to-event distribution, and the other two deep learning based models failed to capture the differences. AFT also remains consistent, but VSI still is the lowest apart from the reference.\par

% put the simulation results here for formatting
\begin{table*}[t!]
\centering
\caption{Model performance summary for simulation study based on $C_{td}$, C-Index and average test log-likelihood. Confidence Intervals for C-Index provided in the SM. For NA entries, the corresponding evaluation metric can not be applied. }
\begin{tabularx}{\textwidth}{c *{9}{Y}}
% \begin{tabular}{@{}llllllllll@{}}
\toprule
Models
 & \multicolumn{3}{c}{$C_{td}$}  
 & \multicolumn{3}{c}{C-Index Raw}
  & \multicolumn{3}{c}{log-likelihood}\\

\cmidrule(lr){2-4} \cmidrule(l){5-7} \cmidrule(l){8-10}
            & 100\%                           & 50\%                            & 30\%                            & 100\%       & 50\%  & 30\%  & 100\%          & 50\%  & 30\%  \\ 
            \midrule
CoxPH       & \textbf{0.757} & 0.755                           & 0.761                           & 0.773       & 0.781 & 0.793 & NA             & NA    & NA    \\
Coxnet      & NA                              & NA                              & NA                              & 0.776       & 0.784 & 0.760 & NA             & NA    & NA    \\
AFT-Weibull & 0.742                           & 0.750                           & 0.768                           & 0.773       & 0.781 & 0.793 & -4.43          & -2.29 & -1.47 \\
RSF         & 0.631                           & 0.638                           & 0.608                           & 0.701       & 0.718 & 0.712 & -14.12         & -8.02 & -5.35 \\
DeepSurv    & NA                              & NA                              & NA                              & 0.772       & 0.781 & 0.793 & NA             & NA    & NA    \\
MLP         & 0.744                           & 0.751                           & 0.770                           & 0.772       & 0.781 & 0.793 & \textbf{-4.15}          & \textbf{-2.22} & -1.41 \\
[5pt]
VSI-NoQ     & 0.748                           & 0.749                           & 0.763                           & 0.772       & 0.781 & 0.793 & -4.16          & \textbf{-2.22} & -1.41 \\
VSI         & 0.748                           & \textbf{0.756} & \textbf{0.772} & 0.773       & 0.781 & 0.793 & \textbf{-4.15}          & \textbf{-2.22} & \textbf{-1.40} \\ \bottomrule
\end{tabularx}
\vspace{-1em}
\label{tab:simubigtable}
\end{table*}

\vspace{3pt}
{\bf Average log-likelihood and C-Index} To validate the effectiveness of VSI, we also provide a comprehensive summary of model performance against other popular or state-of-the-art alternatives in Table~\ref{tab:simubigtable}, under various simulation setups with different evaluation metrics. 
% As shown in Table~\ref{tab:simubigtable}, 
VSI consistently outperforms its counterparts in terms of the average log-likelihood and time-dependent C-Index. Together with the observation that VSI also yields better KS distance (see Table~\ref{Tab:simulationKS}), converging evidence suggests our VSI better predicts the individual survival distributions relative to other competing solutions. 

% With , VSI still performs the best.
We also compared the raw C-Index and the corresponding confidence intervals using the weighted average of model predicted survival time (defined in Sec~\ref{sec:wtaverage}) as the risk score, and we did not find significant differences between alternative methods, as shown in Table~\ref{tab:simubigtable} and Supplemental Materials. Thus VSI can deliver comparable performance relative to models that are compatible with the data generating mechanism. Raw C-Index quantifies the corresponding pairs without considering the time horizon, thus the distinctions among good performing models are not significant.

% \cytao{Maybe we can also present the quantile range [0.05,0.95] of the $\log$-likelihood to show that VSI model likelihoods are tighter.} 

\begin{figure}[t!]
\centering
\begin{tabular}{c@{}c}
\subfloat[]{\includegraphics[width=0.5\linewidth]{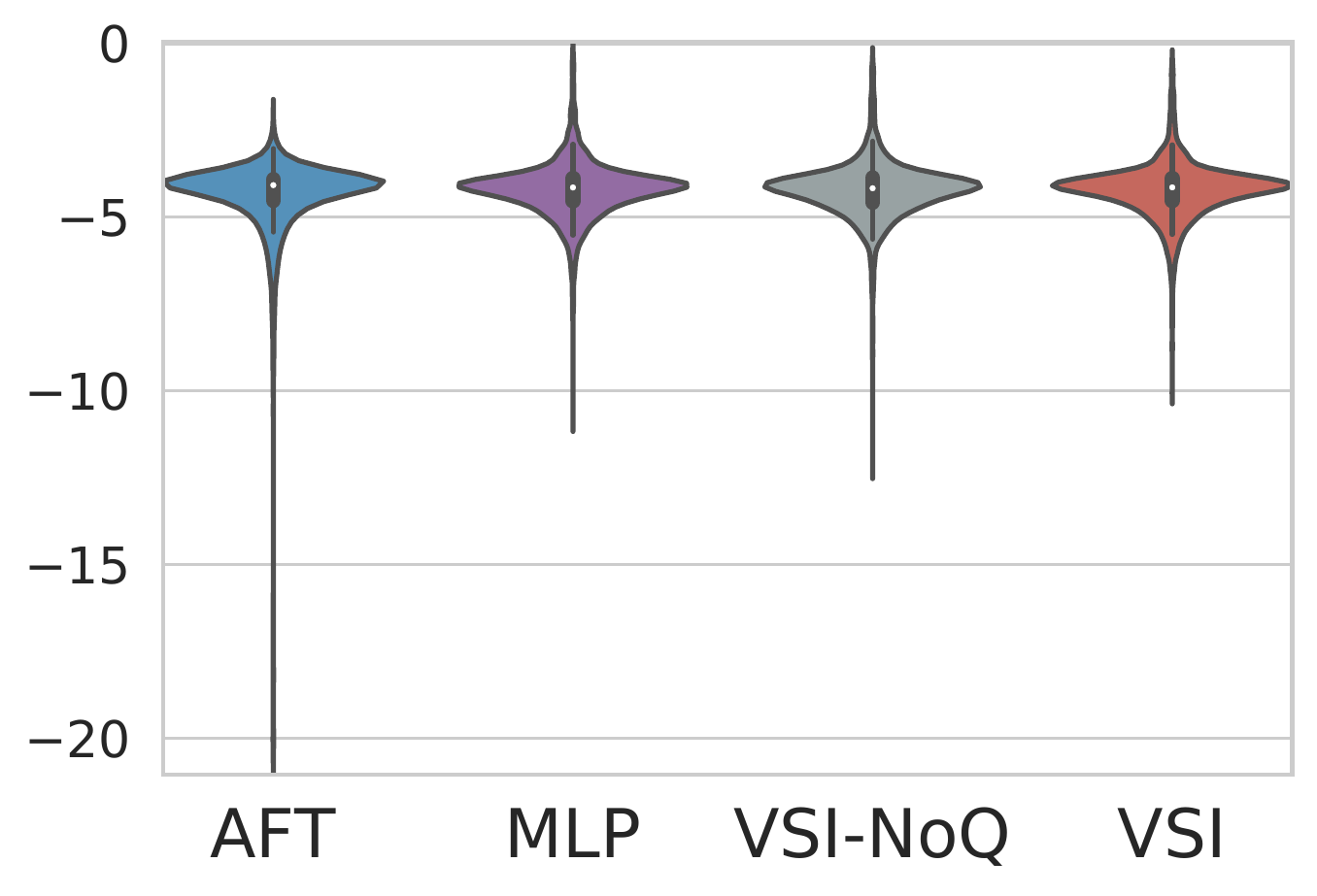}}&
\subfloat[]{\includegraphics[width=0.5\linewidth]{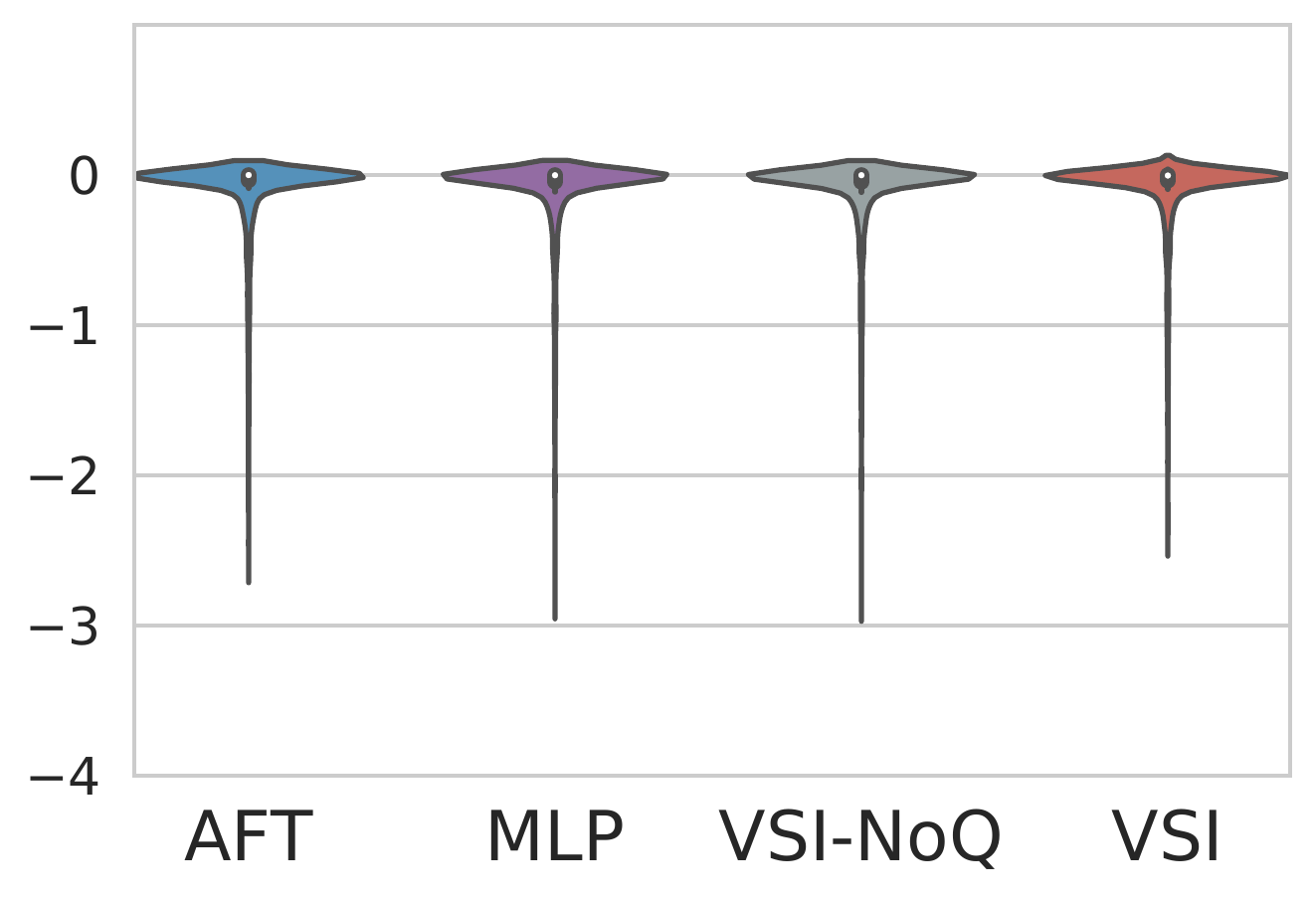}}\\
\end{tabular}
\caption{Test log-likelihood distributions for the 50\% event rate simulation dataset. (left: events, right:censoring)}
\label{fig:simulikeli}
% \vspace{-2em}
\end{figure}

%
%%%% below is [0.05,0.95] range
% \begin{table}[]
% \begin{tabular}{llllll}
% \hline
% Models      & Observed &       &       & Censored &       \\
%             & 100\%    & 50\%  & 30\%  & 50\%     & 30\%  \\ \hline
% AFT-Weibull & 2.548    & 2.618 & 2.861 & 0.557    & 0.542 \\
% MLP         & 2.572    & 2.414 & 2.889 & 0.522    & 0.481 \\
% VSI-NoQ     & 2.584    & 2.327 & 2.841 & 0.519    & 0.492 \\
% VSI         & 2.501    & 2.434 & 2.597 & 0.508    & 0.572 \\ \hline
% \end{tabular}
% \end{table}
%
To provide a more informative summary, We plot the test log-likelihood distributions for selected models in Figure~\ref{fig:simulikeli}. We can see that VSI log-likelihoods estimates are tighter and higher for both observed and censored observations, especially when we have low event rates. The $(0.10, 0.90)$ percentiles range for simulation studies please refer to SM.
\vspace{3pt}

\begin{figure}[t!]
\centering
\begin{tabular}{c@{}c}
\subfloat[]{\includegraphics[width=0.5\linewidth]{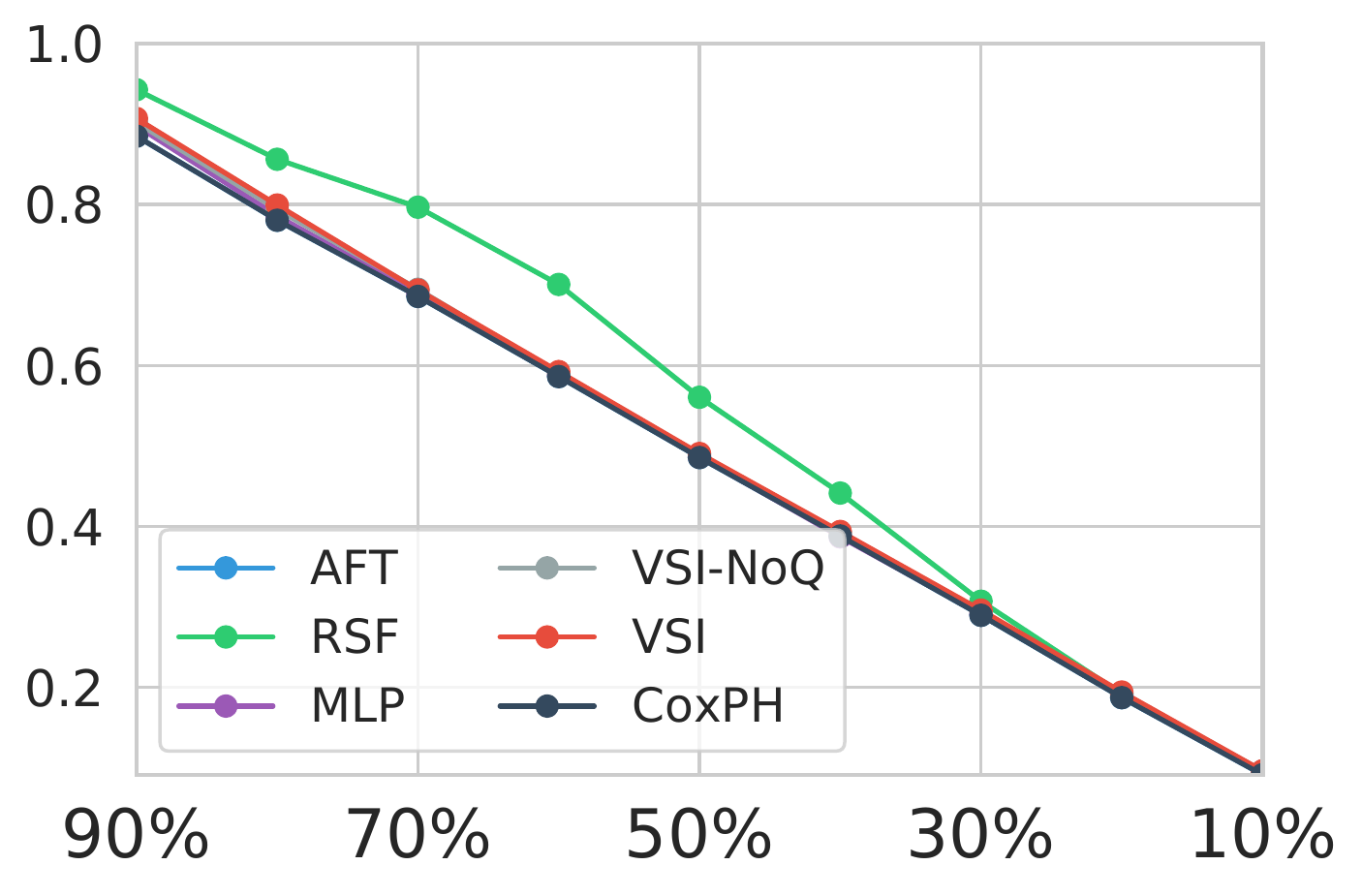}}&
\subfloat[]{\includegraphics[width=0.5\linewidth]{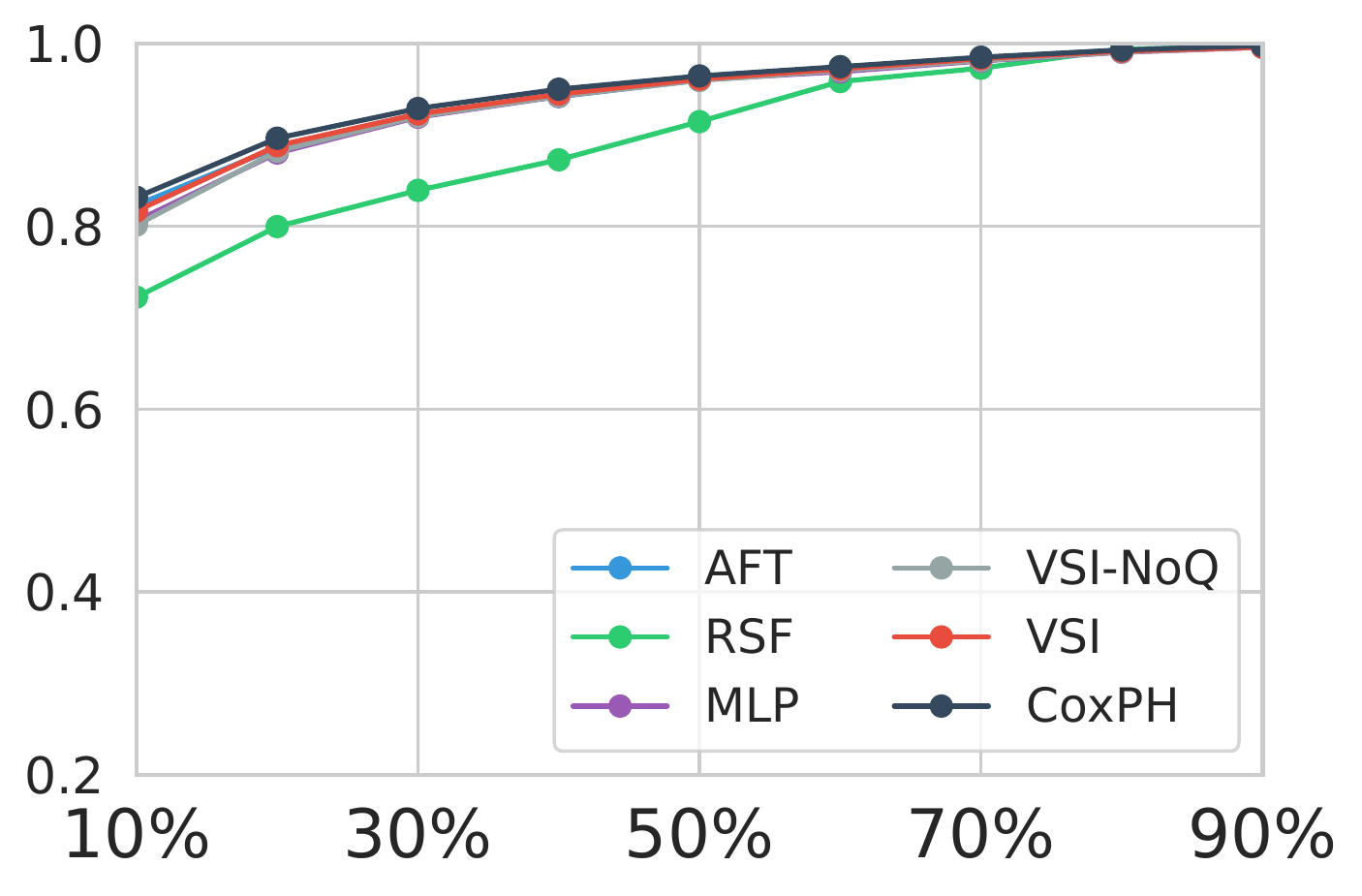}}\\
\end{tabular}
\caption{Test coverage rate for the 50\% event rate simulation dataset. (left: events, right: censoring)}
\label{fig:simucov30}
% \vspace{-2em}
\end{figure}

{\bf Coverage Plots} In Figure~\ref{fig:simucov30}, VSI achieves both relatively high coverage for event (Figure~\ref{fig:simucov30}(a)) and censored observations (Figure~\ref{fig:simucov30}(b)), comparing to the oracle method CoxPH in this synthetic example. Note that while RSF performs better for the observed events, its performance on censored cases falls well below other solutions.

We refer the readers to our Supplementary Materials for additional simulations and analyses based on toy models.
%
% In the Supplementary Materials, we show test log-likelihood distributions showing that VSI offers tighter log-likelihoods for both observed and censored observations. 
% The log-likelihood distribution for different simulation settings among different methods are shown in Figure \ref{fig:simulikeli}, where VSI showed high and concentrated log-likelihood in both events and censoring likelihood.\par
% \vspace{-5pt}
\subsection{Real-World datasets}
% \vspace{-5pt}
Moving beyond toy simulations, we further compare VSI to competing solution on the following three real-world datasets,
$i$) \texttt{FLCHAIN} \citep{dispenzieri2012use}: a public dataset to determine whether the elevation in free light chain assay provides prognostic information to the general population survival,
$ii$) \texttt{SUPPORT} \citep{knaus1995support}: a public dataset for a prospective cohort study to estimate survival over seriously ill hospitalized adults for 180 days period, and
$iii$) \texttt{SEER} \citep{ries2007seer}: a public dataset aim to study cancer survival among adults, which contains 1988 to 2001 information, provided by U.S. Surveillance, Epidemiology, and End Results (SEER) Program.
In this experiments, we used 10-year follow-up breast cancer subcohort in \texttt{SEER} dataset.
We follow the data pre-processing steps outlined in  \citet{chapfuwa2018adversarial}.
% \new{Missing values have been imputed with median (continuous variables) or mode (categorical variables) for simplicity without further assumptions on the original datasets.}
To handle the missing values in data, we adopt the common practice of median imputation for continuous variables and mode imputation for discrete variables. 

\begin{table}[h]
\centering
\caption{Summary Statistics for Real Datasets.}
\begin{tabular}{lrrr}
\toprule
                          & \textsc{FLCHAIN}              & \textsc{SUPPORT}              & \textsc{SEER}                  \\ \midrule
$N$                       & 7,894                & 9,105                & 68,082                \\
Event rate($\%$)          & 27.5                 & 68.1                 & 51.0                  \\
$p$(cat)                  & 26(21)               & 59(31)               & 789(771)              \\
NaN($\%$)                 & 2.1                  & 12.6                 & 23.4                  \\
Max event $t$             & $4998_{\text{days}}$ & $1944_{\text{days}}$ & $120_{\text{months}}$ \\
Loss of  Info($\%$) & 10.45                & 1.57                 & 0.0                   \\ \bottomrule
\end{tabular}
\label{tab:realsummary}
% \vspace{-2em}
\end{table}

Summary statistics of the datasets are shown in Table~\ref{tab:realsummary}, where $N$ is the total number of observations, $p$ denotes the total number of variables after one-hot-encoding, NaN($\%$) stands for the proportion of missingness in covariates, and loss of information stands for the proportion of censoring observations happened after the maximum event time $t$.

%%%%%%for formatting purposes put the real world results here
\begin{table*}[t!]
\centering
\caption{Summary for Real Datasets based on C-Index and average log-likelihood. Confidence Intervals for C-Index are provided in SM. NA implies the corresponding evaluation metric can not be evaluated.}
% NA stands for we could not provide the value for the method. }
\begin{tabularx}{\textwidth}{c *{9}{Y}}
% \begin{tabular}{@{}llllllllll@{}}
\toprule
Models
 & \multicolumn{3}{c}{$C_{td}$}  
 & \multicolumn{3}{c}{C-Index Raw}
  & \multicolumn{3}{c}{log-likelihood}\\

\cmidrule(lr){2-4} \cmidrule(l){5-7} \cmidrule(l){8-10}
            & \textsc{FLCHAIN}                          & \textsc{SUPPORT}                           & \textsc{SEER}                             & \textsc{FLCHAIN}       & \textsc{SUPPORT}   & \textsc{SEER}  & \textsc{FLCHAIN}          & \textsc{SUPPORT}   & \textsc{SEER}  \\ 
            \midrule
 Coxnet      & NA       & NA      & NA    & 0.790       & 0.797   & 0.819 & NA             & NA      & NA    \\
AFT-Weibull & 0.777    & 0.752   & NA    & 0.792       & 0.797   & NA    & -3.09          & -4.39   & NA    \\
RSF         & NA       & NA      & NA    & 0.771       & 0.751   & 0.796 & NA             & NA      & NA    \\
DeepSurv    & NA       & NA      & NA    & 0.785       & 0.678   & NA    & NA             & NA      & NA    \\
MLP         & 0.775    & 0.768   & 0.821 & 0.751       & 0.811   & 0.811 & -1.91          & -2.86   & -2.50 \\
[5pt]
VSI-NoQ     & 0.745    & 0.772   & 0.820 & 0.745       & 0.824   & 0.809 & -2.45          & -2.79   & -2.50 \\
VSI         & \textbf{0.787}    & \textbf{0.775}   & \textbf{0.824} & \textbf{0.792}       & \textbf{0.827}   & \textbf{0.826} & \textbf{-1.85}          & \textbf{-2.74}   & \textbf{-2.49} \\ \bottomrule

\end{tabularx}
\label{tab:realbigtable}
\end{table*}
%%%%%%
%
\begin{table}[ht]
\caption{Quantile ranges for $\log$-likelihood in Real Datasets. Note AFT did not converge to reasonable solutions for SEER. }
% \begin{tabular}{lllllll}
\begin{tabularx}{\columnwidth}{c *{6}{Y}}
\hline
Models      & Observed &         &       & Censored &         &       \\
            & \textsc{flchain}  & \textsc{support} & \textsc{seer}  &  \textsc{flchain}  & \textsc{support} & \textsc{seer}  \\ \hline
AFT & 2.491    & 4.706   & NA    & {\bf 0.468}    & 1.850   & NA    \\
MLP         & 2.970    & 4.273   & 1.780 & 0.518    & 1.540   & 0.623 \\
VSI-NoQ     & 7.34     & 4.744   & 1.801 & 0.559    & 1.634   & 0.529 \\
VSI         & {\bf 2.213}    & {\bf 4.143}   & {\bf 1.718} & 0.537    & {\bf 1.354}   & {\bf 0.508} \\ \hline
\end{tabularx}
\label{tab:realqrangelikeli}
\end{table}
In Table~\ref{tab:realbigtable} we compare the C-Indices and average log-likelihood. The advantage of VSI is more evident for the more challenging real datasets, especially in the cases of low observed event rates. For example, with 30\% event rate, in \textsc{SUPPORT} dataset, VSI Confidence Interval for raw C-Index as (0.809, 0.846), while the standard CoxNet is only (0.763,0.805) and AFT (0.782,813), {\it i.e.}, the overlaps with that of VSI are very small. Similar results were observed for other datasets and baseline solutions. VSI shows remarkable robustness against data incompleteness in a real-world scenario, achieving the best results according to all three metrics. For VSI the raw C-Index is computed from the weighted average of VSI predicted distribution, please refer to SM for more details.
%  highest C-Index and $\mathcal{C}_{td}$, and log-likelihood
%
\begin{figure}[ht]
\centering
\begin{tabular}{c@{}c}
\subfloat[]{\includegraphics[width=0.50\linewidth]{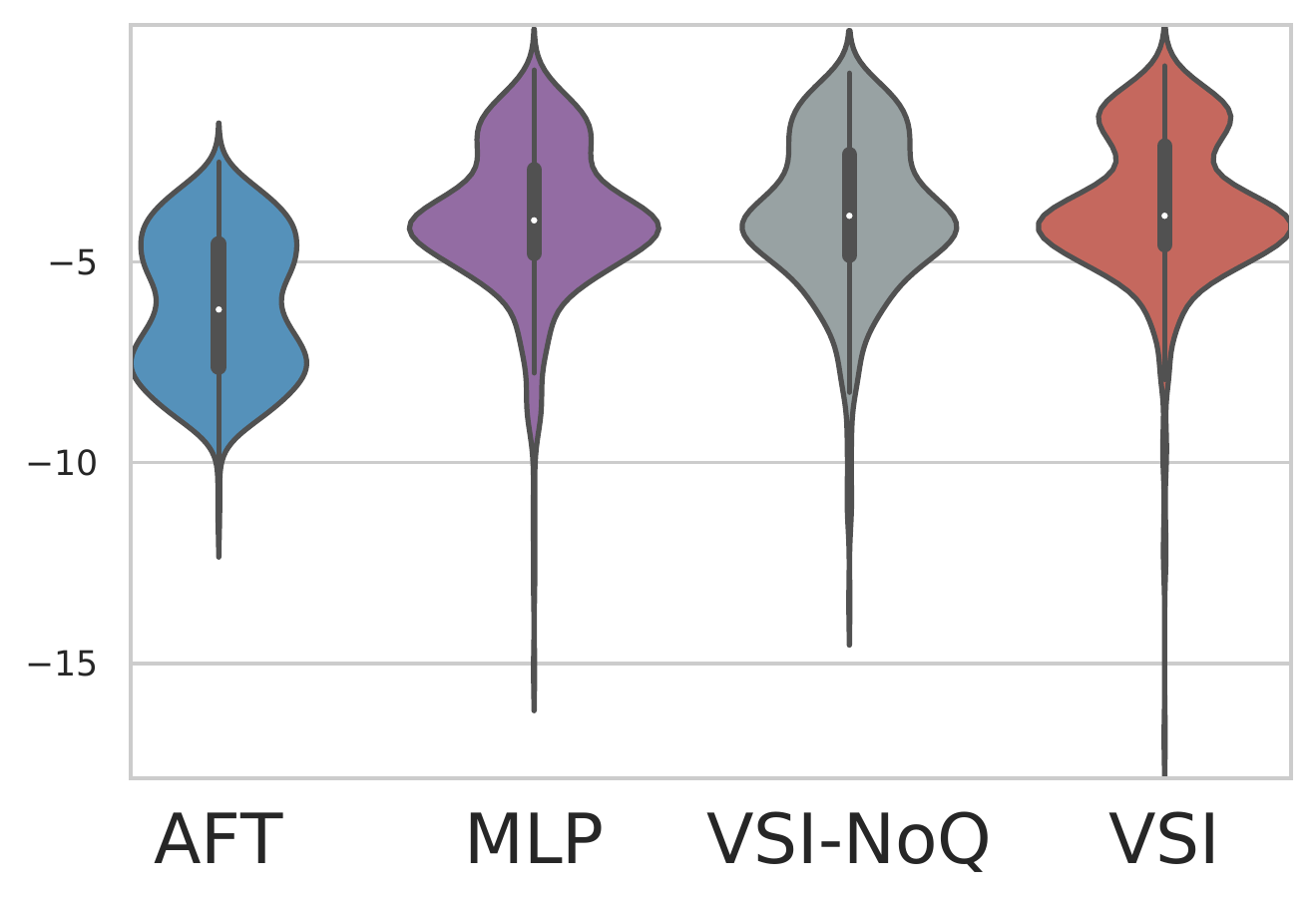}} &
\subfloat[]{\includegraphics[width=0.50\linewidth]{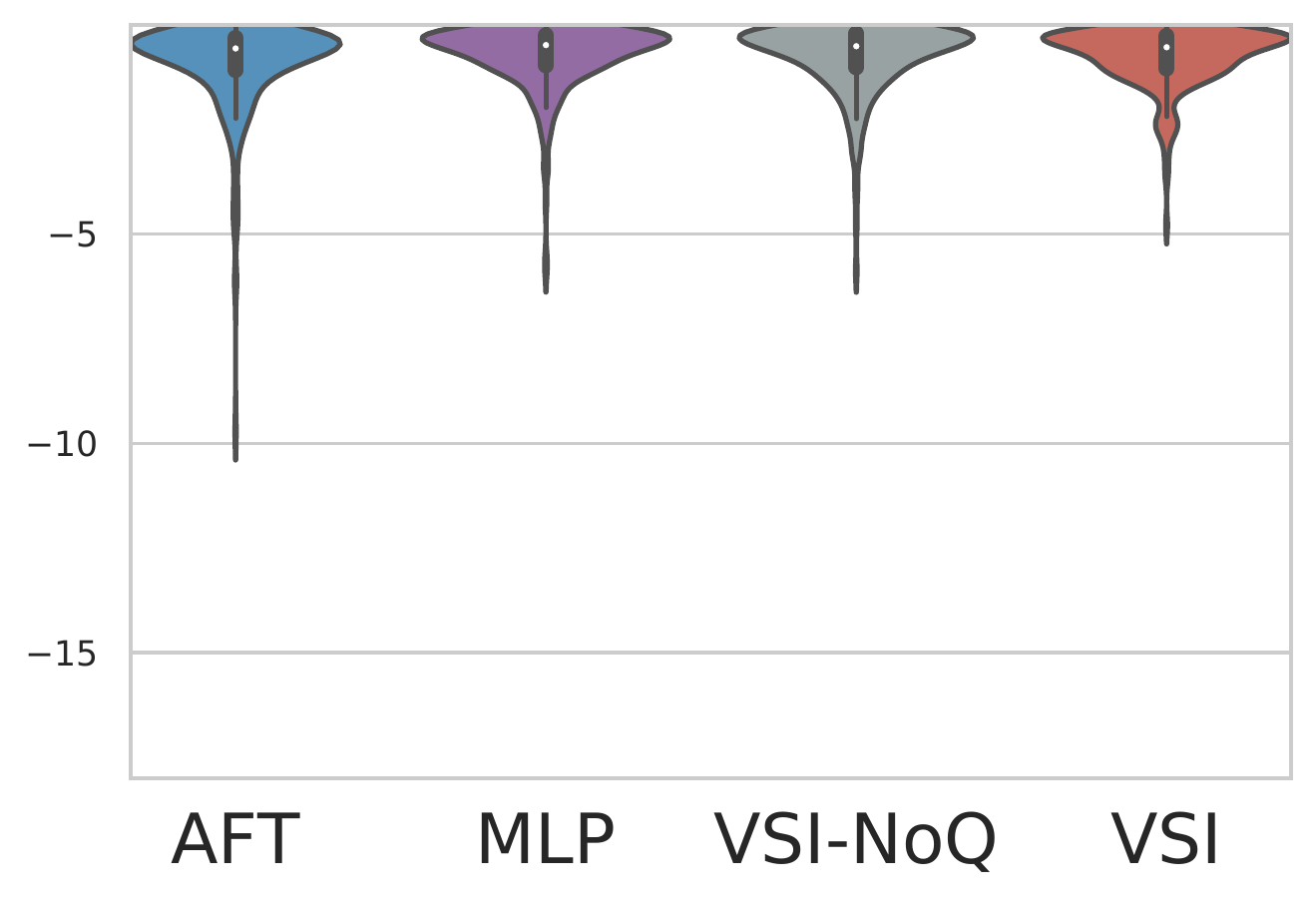}}
\end{tabular}
\caption{$\log$-likelihood distributions for \texttt{SUPPORT} Dataset, (left: events, right:censoring)}
\label{fig:reallikeli}
% \vspace{-1em}
\end{figure}
In Figure \ref{fig:reallikeli}, the distribution of $\log$-likelihood is more concentrated, in addition to a higher mean. To quantitatively evaluate the concentration, we report the difference between $10\%$ and $90\%$ quantiles of $\log$-likelihood in Table~\ref{tab:realqrangelikeli}. The quantile ranges of VSI are considerably smaller compared to alternative solutions under most experimental settings. This verifies VSI enjoys better model robustness compared to other popular alternatives, especially in the case of high censoring rates. 
% \new{}
%
\begin{figure}[ht]
\centering
\begin{tabular}{c@{}c}
\subfloat[]{\includegraphics[width=0.50\linewidth]{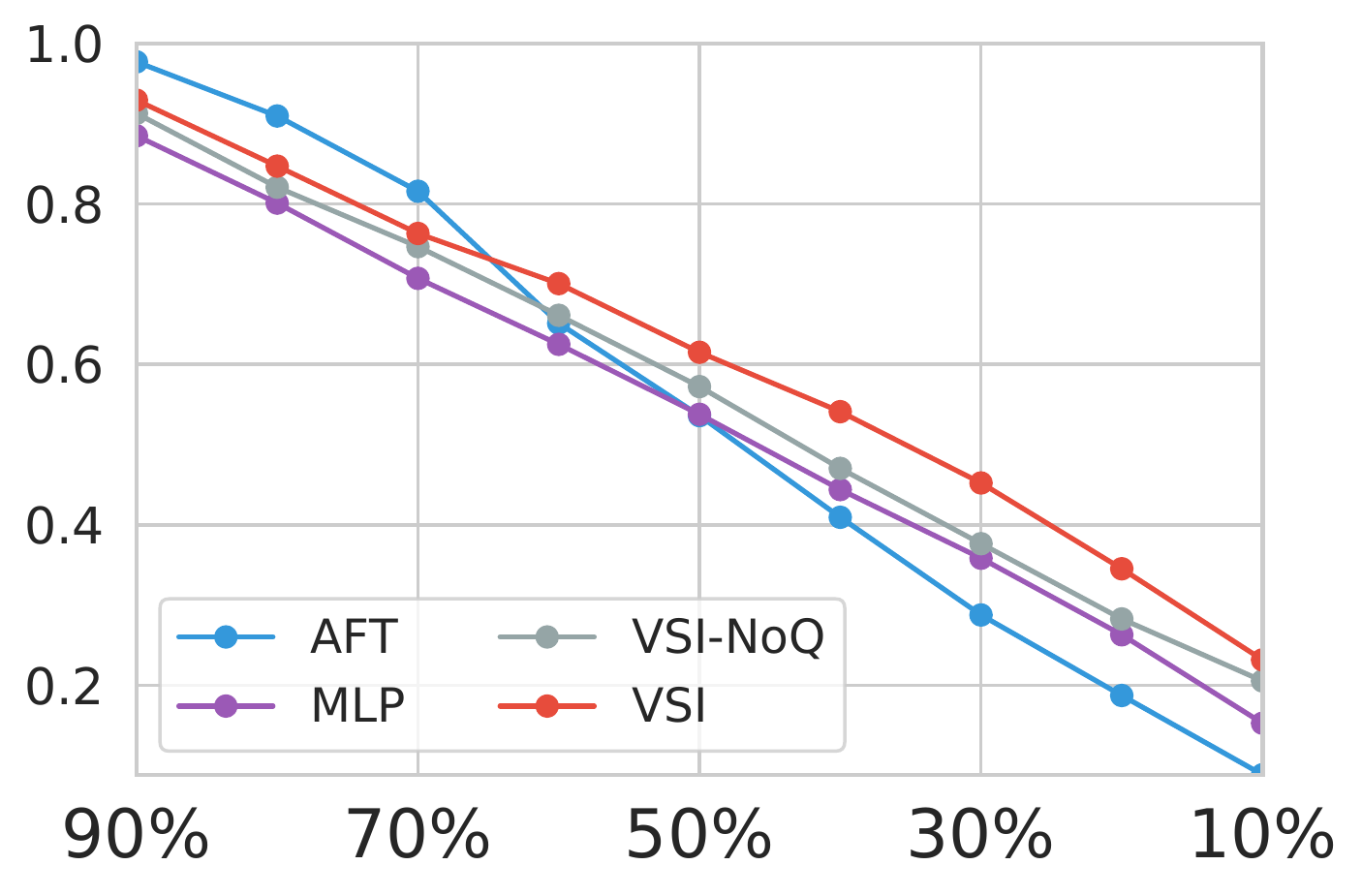}} &
\subfloat[]{\includegraphics[width=0.50\linewidth]{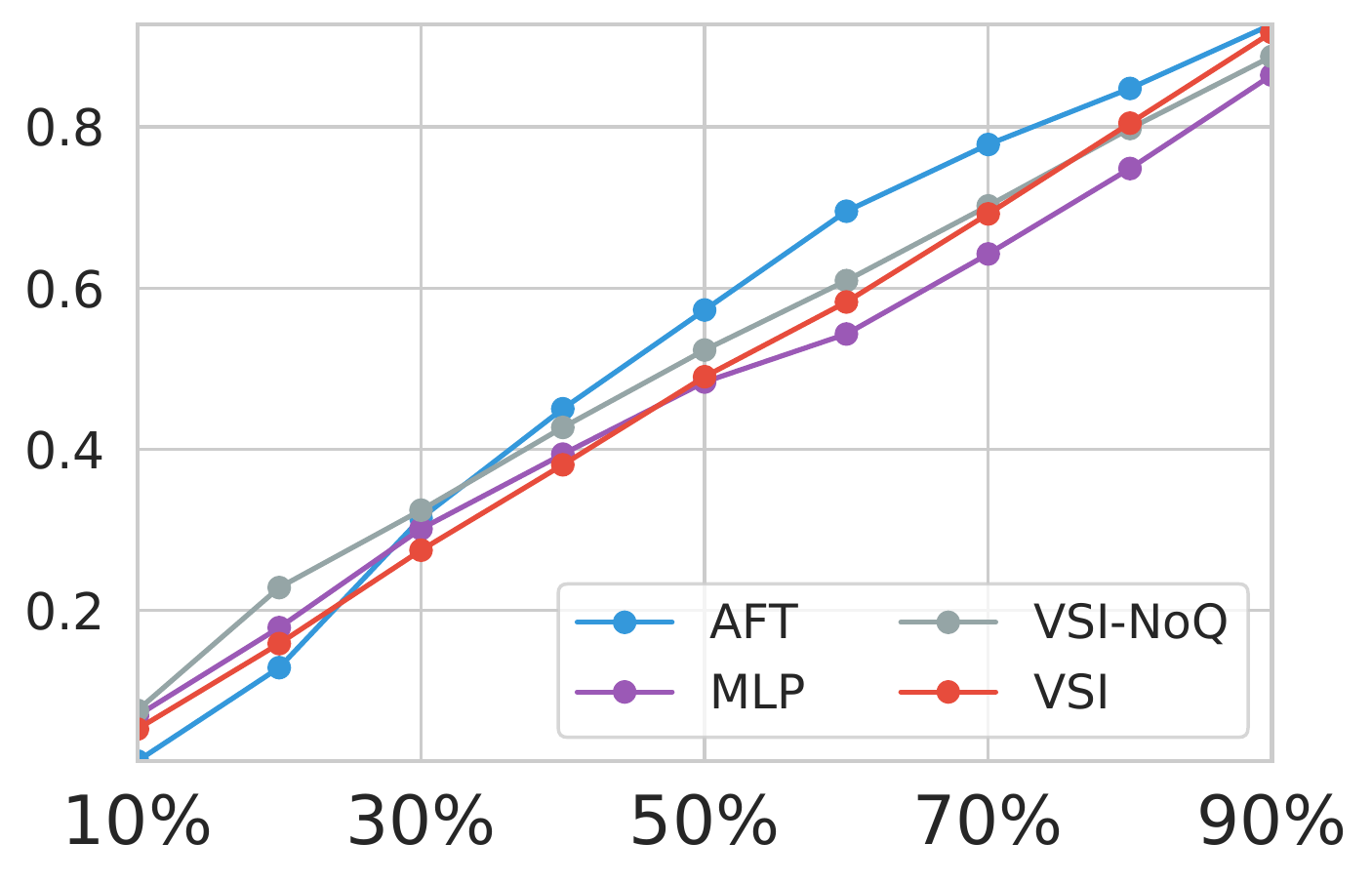}}
\end{tabular}
\caption{Coverage rate for \texttt{SUPPORT} Dataset, (left: events, right: censoring)}
\label{fig:realcovsupport}
\vspace{-1em}
\end{figure}

Together with the coverage plots in Figure \ref{fig:realcovsupport}, VSI has relative high coverage for both events and censoring cases which indicates better performance in capturing the true event time in challenging real-world datasets. The consistency of those results have been verified through repeated runs on these three datasets. For more detailed results please refer to SM.

% \vspace{-5pt}
\section{CONCLUSIONS}
% \vspace{-5pt}
%
We presented an approach for learning time-to-event distributions conditioned on covariates in a nonparametric fashion by leveraging a principled variational inference formulation.
The proposed approach, VSI, extends the variational inference framework to survival data with censored observations.
Based on synthetic and diverse real-world datasets, we demonstrated the ability of VSI to recover the underlying unobserved time-to-event distribution, as well as providing point estimations of time-to-event for subjects that yield excellent performance metrics consistently outperforming feed-forward deep learning models and traditional statistical models.

As future work, we plan to extend our VSI framework to longitudinal studies, where we can employ a recurrent neural net (RNN) to account for the temporal dependencies. For datasets with observations made at irregular  intervals, for instance, the Neural-ODE model \citep{chen2018neural} can be applied. Our work can be also adapted to make dynamic predictions of event times to serve the needs of modern clinical practices.

\subsubsection*{Acknowledgements}
The authors would like to thank the anonymous reviewers for their insightful comments. This research was supported in part by by NIH/NIBIB R01-EB025020. 

% Use the unnumbered third level heading for the acknowledgements.  All
% acknowledgements go at the end of the paper.

% \clearpage
\bibliographystyle{ACM-Reference-Format}
\bibliography{VSI-arXiv}

\clearpage

\beginsupplement

\appendix
\section{Implementing VSI}
For model inputs, we encode each subject's time with a one-hot vector $t^\text{OH}$ with dimension $(1 \times M)$ based on the observed event $t_i$ for subject $(\delta_i=1)$. With event $t$ has been one-hot-encoded, only one bin where the event time falls in equals 1, else equal 0. For censoring observations $(\delta_i=0)$ we have partial information about the true unknown event time, instead of missingness. We know for sure that the events would happen after the observed censored time $t_i$. The input vector, timepoints after the censored time $t_i$ have been put a prior based on population time-to-event distribution by Nelson-Aalen Estimator. This re-weighting strategy informs the model that this subject didn't have event before time $t_i$, and also regularizes the model with respect to empirical tail distribution. One may also choose not to encode $t_i$ but adding $\delta_i$ to the input as well. In this case, each subject in the training has input of the form $(x_i,t_i,\delta_i)$. To predict new $t$, the input for the prediction process is $(x_i, \delta=1)$ where by defaut we are predicting the time-to-event. We have shown by experiments the two form of inputs produce similar results and the encoding strategy performs better under both simulation and real-world scenarios. The KS statistic for 30\% event rates has 0.073, well VSI has 0.059.

\vspace{3pt}
{\bf Population Survival Estimation Based on Nelson-Aalen Estimator} For NA estimator, $\tilde{H}(t) = \sum_{t_j\le t} \frac{d_j}{n_j}$. Therefore we could calculate $f(t)$ accordingly. Let the selected discretized time be $t_{(b)}$, where $b\in \{1,\ldots, M\}$.
\begin{equation*}
\begin{split}
\tilde{H}(t_{(b)}) &= \sum_{j\le b} \frac{d_j}{n_j}\\
\tilde{S}(t_{(b)}) &= \exp(-\tilde{H}(t_{(b)})) \\
\tilde{f}(t_{(1)}) &= 1-\tilde{S}(t_{(1)}) \\
\tilde{f}(t_{(b)}) &= \tilde{S}(t_{(b)}) - \tilde{S}(t_{(b-1)}) \\
&=  \exp(- \sum_{j\le b} \frac{d_j}{n_j}) -  \exp(- \sum_{j\le b-1} \frac{d_j}{n_j})\\
\tilde{f}(t_{(M)}) &= 1-\sum_{b=1}^{M-1}\tilde{f}(t_{(b)})
\end{split}
\end{equation*}
Therefore the we have $[\tilde{f}(t_{(1)}), \tilde{f}(t_{(2)}),\ldots,\tilde{f}(t_{(M)})]$ accordingly for each timepoint. For each censored subject where the censored time falls in bin $k$, 

\[
    t^\text{OH}_{(b)} = 
\begin{cases}
    0,& \text{if } b\le k\\
    \frac{\tilde{f}(t_{(b)}) }{\sum_{b>k}\tilde{f}(t_{(b)}) },              & \text{if } b> k
\end{cases}
\]

In this way the constructed $\sum_b t^\text{OH}_{(b)} =1$ as well.  Now we have the soft one-hot-encoding for original censored $t$. 

{\bf Architecture for VSI}
The input of VSI is $Y_i = (x_i, \delta_i, t^\text{OH}_i)$. Parameters for $p_{\theta}(z|x)$
and $q_{\phi}(z|t^\text{OH}_i,x)$ were obtained by two MLP framework respectively, each with hidden dimensions $[32, 32]$ with output dimension $[32]$.  The decoding arm $p_{\theta}(t_i|z)$ has hidden dimensions $[32, 32, 32]$ with output a $(1 \times M)$ vector which is the predicted logits. Activation function is leaky ReLU, continuous variables are z-transformed before entering the model based on training datasets.

\section{Derivation of Likelihood lowerbound for censored observations}
To get the likelihood lowerbound for censored observations, we applied Fubini's  theorem and Jensen's inequality.
\begin{equation*}
\begin{split}
& \mathcal{L}_c(x_i, t_i;\theta)\\
&=\text{log }S_{\theta}(t_i|x_i)\\&= \text{log } \int_{t_i}^{\infty}p_{\theta}(t|x_i) dt\\
&= \text{log } \int_{t_i}^{\infty}\int_z p_{\theta}(t,z|x_i) dzdt\\
&= \text{log } \int_{t_i}^{\infty}\int_z \frac{p_{\theta}(z, t|x_i)}{q_{\phi}(z|t_i,x_i)} q_{\phi}(z|t_i,x_i) dzdt\\
&= \text{log } \int_{t_i}^{\infty}\int_{q_{\phi}(z|t_i,x_i)} \frac{p_{\theta}(z, t|x_i)}{q_{\phi}(z|t_i,x_i)}  dQ(z|t_i,x_i)dt\\
&= \text{log } \int_{q_{\phi}(z|t_i,x_i)} \int_{t_i}^{\infty}\frac{p_{\theta}(z, t|x_i)}{q_{\phi}(z|t_i,x_i)} dt dQ(z|t_i,x_i)\\
&= \text{log } \int_{q_{\phi}(z|t_i,x_i)} \int_{t_i}^{\infty}p_{\theta}(t|z)\frac{p_{\theta}(z|x_i)}{q_{\phi}(z|t_i,x_i)} dt dQ(z|t_i,x_i)\\
&= \text{log } \int_{q_{\phi}(z|t_i,x_i)} \frac{p_{\theta}(z|x_i)}{q_{\phi}(z|t_i,x_i)}\int_{t_i}^{\infty}p_{\theta}(t|z) dt dQ(z|t_i,x_i)\\
&\ge \int_{q_{\phi}(z|t_i,x_i)} \text{log } \left(\frac{p_{\theta}(z|x_i)}{q_{\phi}(z|t_i,x_i)}\int_{t_i}^{\infty}p_{\theta}(t|z) dt \right)dQ(z|t_i,x_i)\\
&=  \mathbb{E}_{q_{\phi}(z|t,x)}[\text{log }{\frac{p_{\theta}(z|x_i)}{q_{\phi}(z|t_i,x_i)}} + \text{log }S_{\theta}(z, t_i|x_i)] \\
&=L_c(x_i,t_i; \theta,\phi)
\end{split}
\label{eq:censorVAElb}
\end{equation*}

{\bf Likelihood for Multivariate Normal Distribution} The Multivariate Normal Log-likelihood function used for ELBO has $z_l\sim N(\mu, \Sigma)$ , where $z_l$ is $m\times 1$, $\mu$ is $m \times 1$ and $\Sigma$ is $m\times m$ :
$$\text{log }\mathcal{L}(z_l;\mu,\Sigma) = -\frac{1}{2}[\text{log }(|\Sigma|) + (z_l-\mu)^T\Sigma^{-1}(z_l-\mu) + m\text{log }(2\pi)]$$

\section{Likelihood Calculation for Statistical Models}

{\bf AFT Models}
Distributions of $T_i$ and $\epsilon_i$ are corresponding to each other, with:
\begin{equation}
\begin{split}
S_i(t|x_i) &= S_{\epsilon_i}(z_i) = S_0(\exp(\beta^Tx_i)t)\\
z_i &= \frac{log(t)-\mu-\theta^Tx_i}{\sigma}\\
h_i(t|x_i) &= \frac{1}{\sigma t_i}h_{\epsilon_i}(z_i) = h_0(\exp(\beta^Tx_i)t)
\end{split}
\end{equation}
The likelihood function under general non-informative censoring has the form:
\begin{equation}
\begin{split}
\mathcal{L}(\mathcal{D};\theta) &= \prod_{i=1}^n h(t_i|x_i)^{\delta_i}S(t_i|x_i)\\
&= \prod_{i=1}^n (\sigma t_i)^{-\delta_i}\{h_{\epsilon_i}(z_i)\}^{\delta_i}S_{\epsilon_i}(z_i)
\end{split}
\end{equation}
For AFT model with Weibull distribution, $\epsilon_i$ follows the extreme value distribution and $T_i$ follows the Weibull distribution, with scale parameter $\lambda>0$, shape parameter $\nu >0$.
\begin{equation*}
\begin{split}
f_0(t) &= \lambda \nu t^{\nu-1} \exp(-\lambda t^{\nu})\\
h_0(t) &= \lambda \nu t^{\nu-1}\\
S_0(t)&=\exp(-\lambda t^{\nu})\\
\E(T) &= \frac{1}{\lambda^{1/\nu}}\Gamma(\frac{1}{\nu}+1)\\
S_{\epsilon}(z) &= \exp(-\exp(z))\\
f_{\epsilon}(z) &= \exp(z)\exp(-\exp(z))\\
h_{\epsilon}(z) &= \frac{f_{\epsilon}(z)}{S_{\epsilon}(z) }\\
&= \exp(z)
\end{split}
\end{equation*}
Thus $\lambda = \exp(-\mu/\sigma)$ and $\nu=1/\sigma$ in the extreme value distribution $\epsilon$, equivalently $\mu = -\frac{\log \lambda}{\nu}$ and $\sigma=1/\nu$. Thus the log-likelihood is:
\begin{equation}
\begin{split}
&\log \mathcal{L}_{\text{Weibull}}(\mathcal{D};\theta) \\
&= \sum_{i=1}^n \left[-\delta_i\log(\sigma t_i) +\delta_i \log\{h_{\epsilon_i}(z_i)\} + \log S_{\epsilon_i}(z_i)\right]\\
&= \sum_{i=1}^n \left[\delta_i\log (\nu) - \delta_i\log( t_i) +\delta_i z_i -\exp(z_i)\right]
\end{split}
\end{equation}
% When we add the covariates, Cox-Weibull is:
% \begin{equation*}
% \begin{split}
% h(t|x) &= h_0(t)\exp(\beta^Tx)=\lambda \nu t^{\nu-1}\exp(\beta^Tx)
% \end{split}
% \end{equation*}
% For log-normal distribution, $S(t;x) = S_0(\exp(-\alpha^Tx)t)=1-\Phi(\frac{\text{log}t-\alpha^Tx - \mu}{\sigma})$.  Thus,
% \begin{equation*}
% \begin{split}
% S(t;x)&=1-\Phi(\frac{\text{log}t- \alpha^Tx}{\sigma(y)})\\
% f(t;x) &= \phi(\frac{\text{log}t- \alpha^Tx}{\sigma(y)})\frac{1}{\sigma(y)t}\\
% H(t;x)&=-\text{log}(1-\Phi(\frac{\text{log}(T)-\alpha^Tx}{\sigma(y)}))\\
% h(t;x)&= \frac{f(t;x)}{S(t;x)}=-\frac{d}{dt}\log(S(t))\\
% % &= \frac{1}{1-\Phi(\frac{\text{log}(T)-\alpha^Tx}{\sigma(y)})}\phi(\frac{\text{log}(T)-\alpha^Tx}{\sigma(y)})\frac{1}{T}\\
% \log \mathcal{L} &= \log \prod h(t_i;x_i)^{\delta_i}S_i(t_i|x_i)\\
% &= \sum \delta_i \log h(t_i;x_i) + \log S_i(t_i|x_i)\\
% &= \sum\delta_i [\log\phi(\frac{\text{log}t- \alpha^Tx}{\sigma(y)})-\log\sigma(y)-\log(t) - \log(1-\Phi(\frac{\text{log}t-\alpha^Tx}{\sigma(y)}))] + \log(1-\Phi(\frac{\text{log}t- \alpha^Tx}{\sigma(y)}))
% \end{split}
% \end{equation*}
\par
Note that for comparison log-likelihood at same scale, AFT and other parametric methods' log-likelihood were calculated at discretized event time same as VSI.

{\bf Random Survival Forest}\par
Random Survival Forest could give estimates for cumulative hazard function at the pre-specifed timepoints $\hat{H}(t_{(b)}), b\in \{1, ldots, M\}$, which could let us get the estimated survival function and time-to-event distribution. 
\begin{equation*}
\begin{split}
\hat{S}(t_{(b)}) &= \exp(-\hat{H}(t_{(b)})) \\
\hat{f}(t_{(1)}) &= 1-\hat{S}(t_{(1)}) \\
\hat{f}(t_{(b)}) &= \hat{S}(t_{(b)}) - \hat{S}(t_{(b-1)}) \\
\hat{f}(t_{(M)}) &= 1-\sum_{b=1}^{M-1}\hat{f}(t_{(b)})\\
\end{split}
\end{equation*}

Then based on the survival likelihood function $$\mathcal{L}(\mathcal{D};\theta) = \prod_{i=1}^n \hat{p}(t_i|x_i)^{\delta_i}\hat{S}(t_i|x_i)^{1-\delta_i}$$, we could get the likelihood for RSF.

{\bf Cox Models}\par
 Since cox model is semiparametric, we don't know the full parameters, therefore usually we would give partial likelihood instead of full likelihood.
\begin{equation*}
\begin{split}
PL(\beta) = \prod_{i=1}^n \{\frac{\exp(\beta^Tx_i)}{\sum_{l\in R(t_{(i)})} \exp(\beta^Tx_i)}\}
\end{split}
\end{equation*}
$R(t_{(i)})$ is the set of subjects at risk at time $t_{(i)}$. Above partial likelihood is from multiply the probability that a subject with covariates $x_i$ dies in $(t_{(i)}, t_{(i)}+\delta t]$ with $\delta t \rightarrow 0$ for each subject with event. $\hat{\beta}$ is obtained by Maximize $\log PL(\beta)$. \par
To make comparison between difference method, we would use some empirical way to calculate the full likelihood for Cox models.
\begin{equation*}
\begin{split}
\mathcal{L}(\mathcal{D};\theta) &= \prod_{i=1}^n h(t_i|x_i)^{\delta_i}S(t_i|x_i)\\
\log \mathcal{L}(\mathcal{D};\theta) &= \sum_{i=1}^n \delta_i [\log h_0(t_i) +\beta^Tx_i]+ \log S(t_i|x_i)
\end{split}
\end{equation*}
In above equations, we would use the estimated $h_0(t)$ and predicted $S(t_i|x_i)$. \\

\section{Additional Results for Simulation Datasets}

\begin{figure}[ht]
\centering
\begin{tabular}{c@{}c}
\subfloat[]{\includegraphics[width=0.49\linewidth]{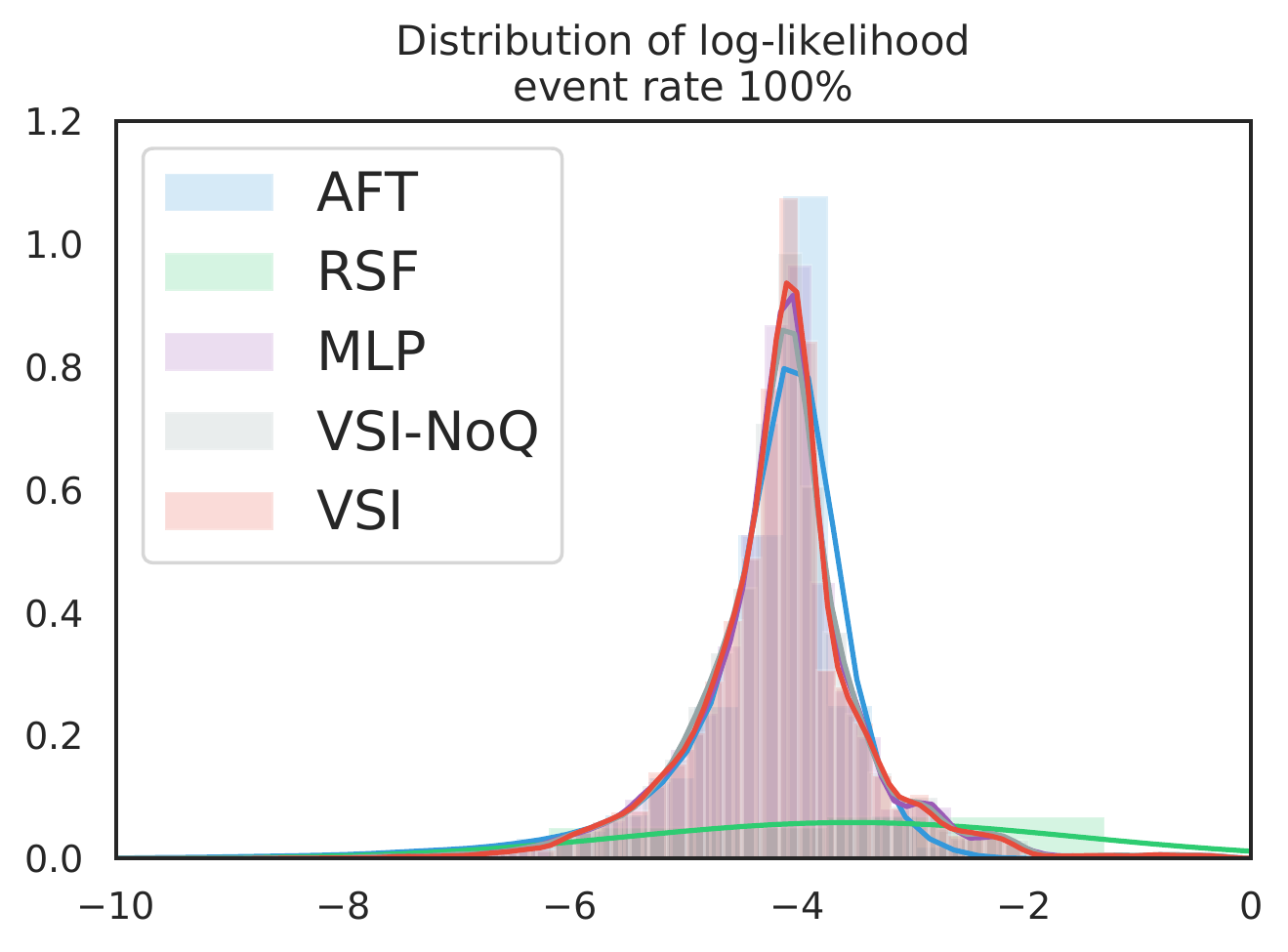}} &\\
 \subfloat[]{\includegraphics[width=0.49\linewidth]{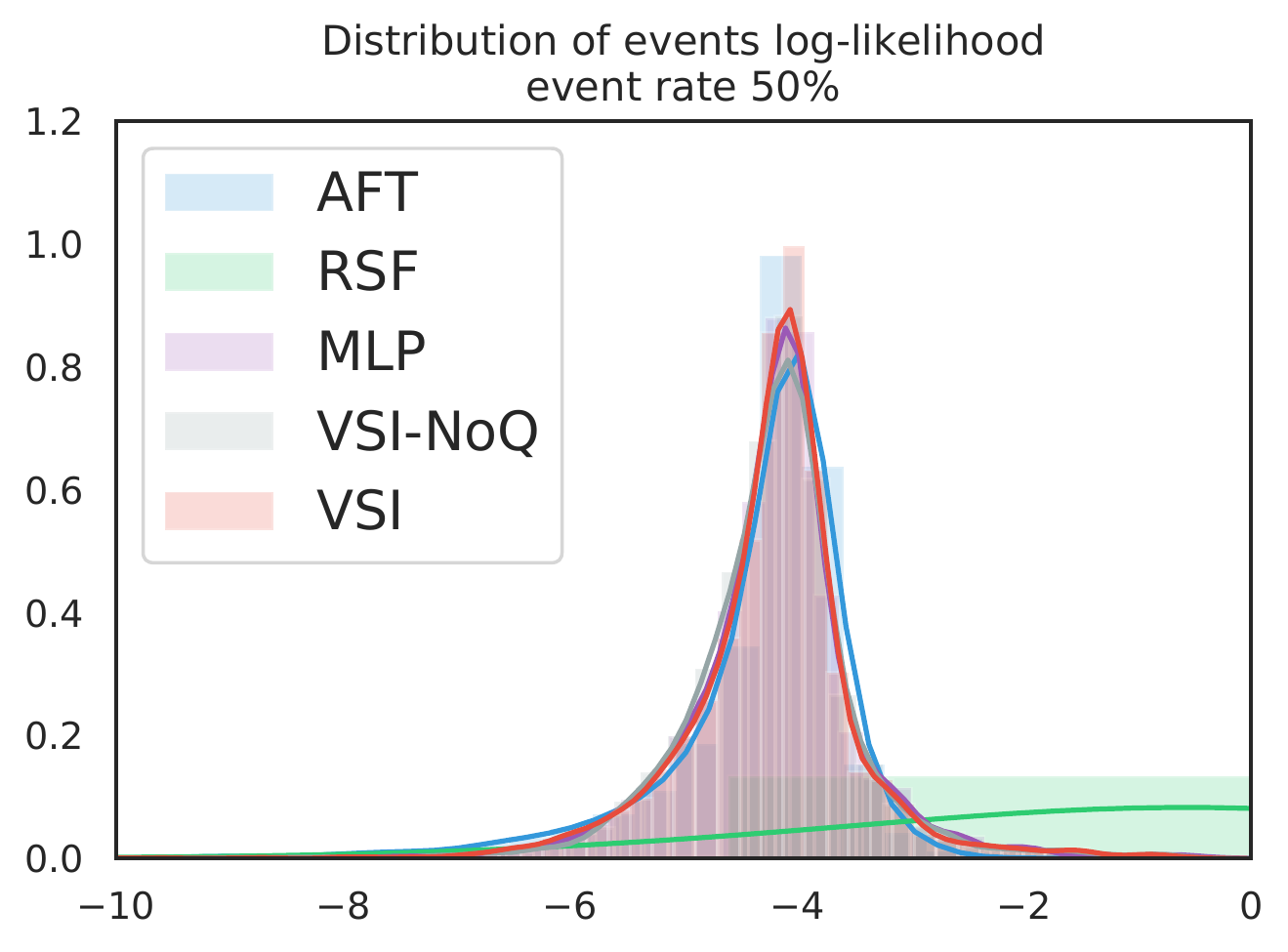}} &  \subfloat[]{\includegraphics[width=0.49\linewidth]{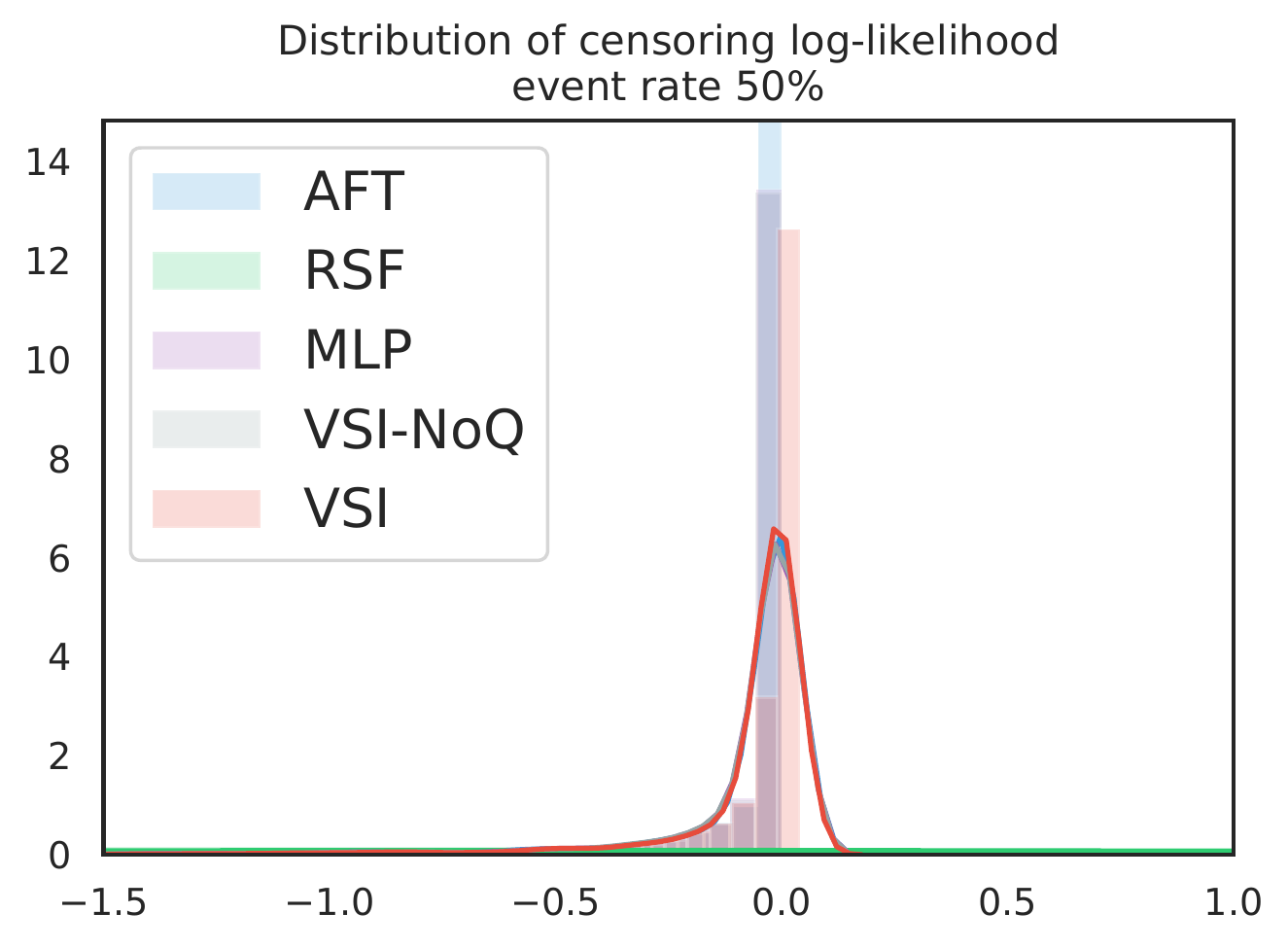}}\\
\subfloat[]{\includegraphics[width=0.49\linewidth]{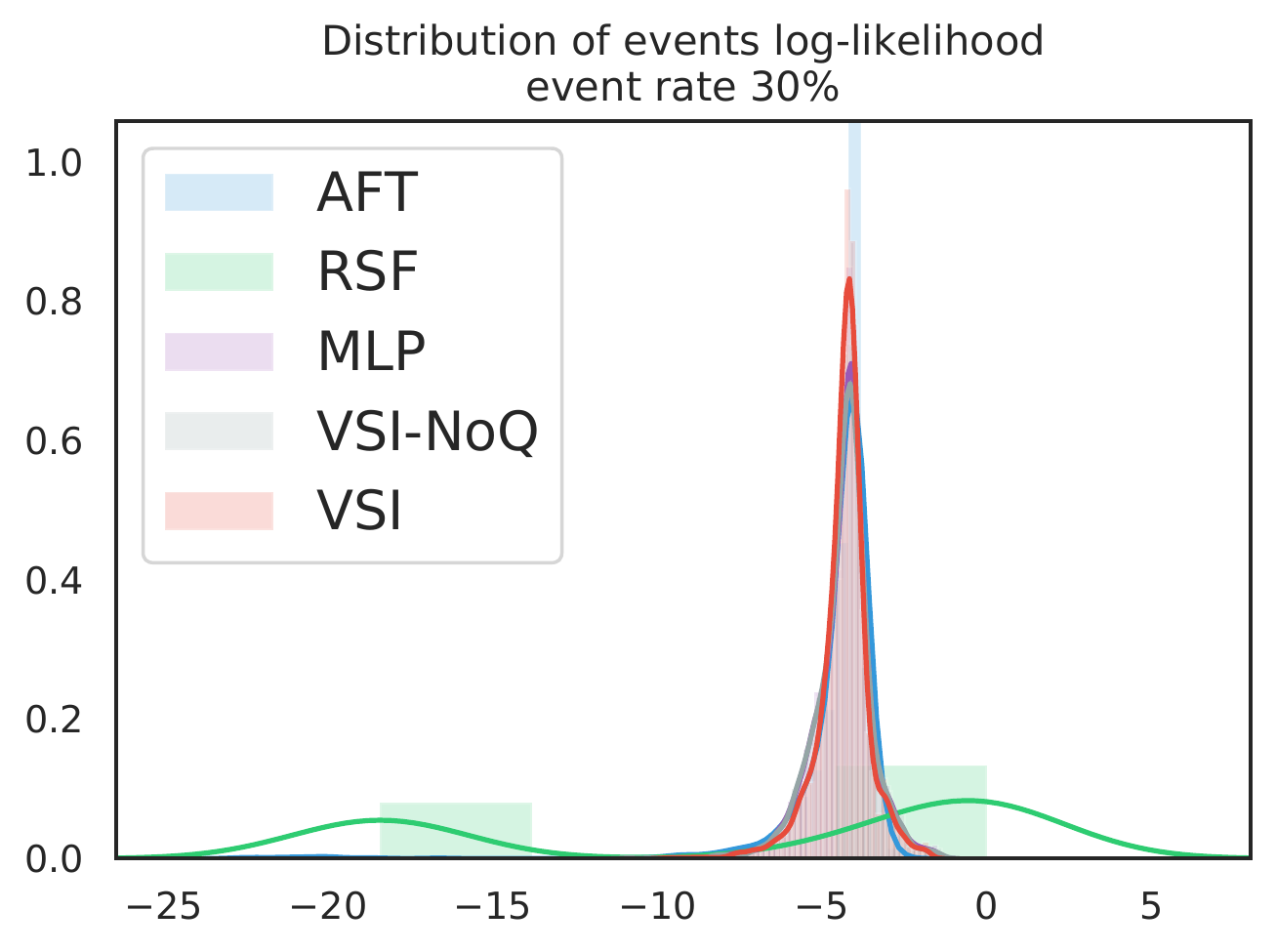}}&
\subfloat[]{\includegraphics[width=0.49\linewidth]{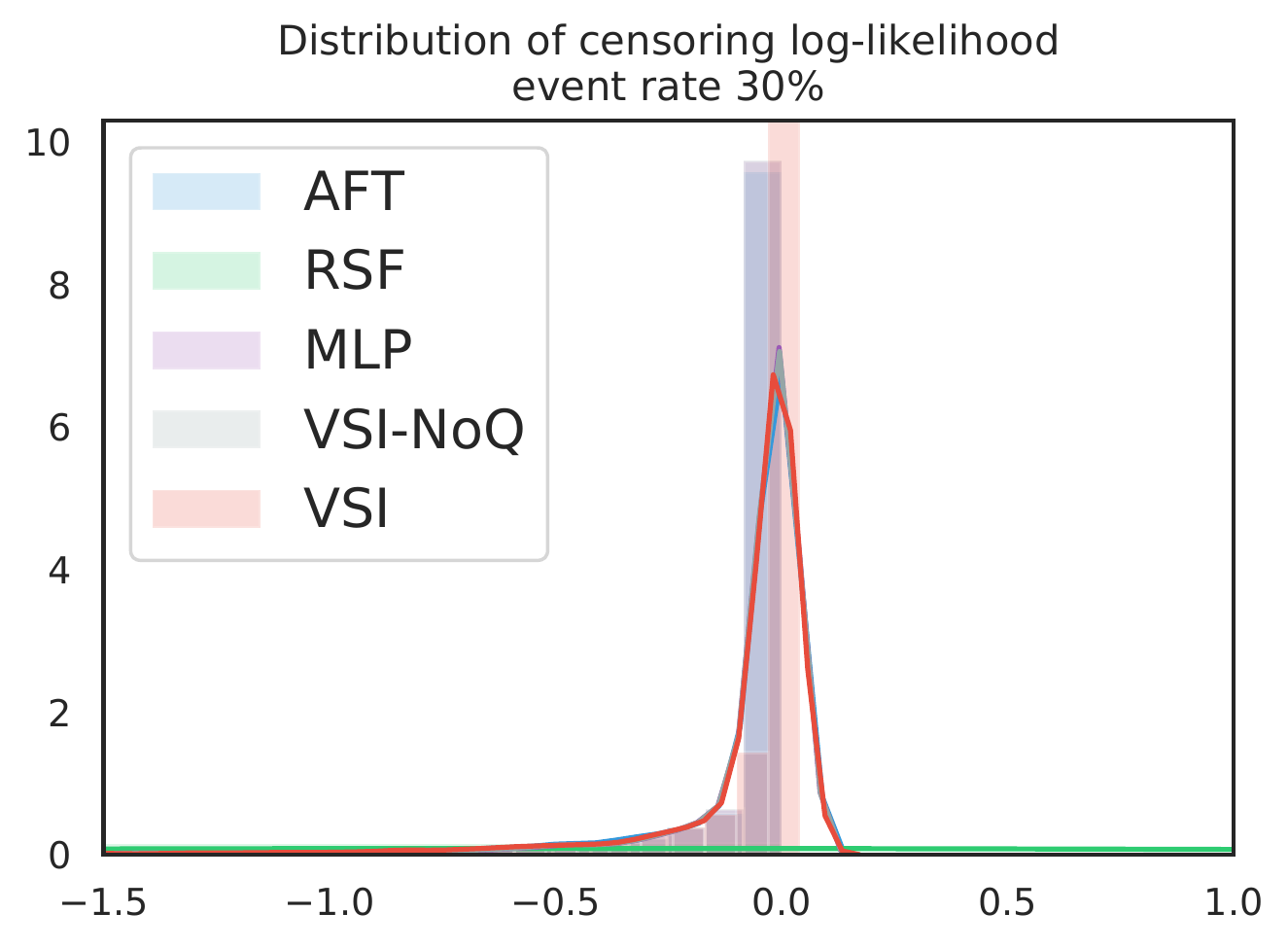}}\\
\end{tabular}
\caption{Testing datasets log-likelihood distributions for simulation dataset. $100\%$ event rate (a), $50\%$ event rate (observed:b, censored:c), $30\%$ event rate (observed:d, censored:e)}
\label{fig:simulikeli-all}
\end{figure}

{\bf Distribution of log-likelihood} The distributions of log-likelihoods for simulation datasets are shown in 
Figure~\ref{fig:simulikeli-all}. In all scenarios, VSI has the most concentrated distribution of the log-likelihood, which suggests that VSI is robust under different event rate with CoxPH assumptions. MLP and VSI-NoQ performed equally good in this simple synthetic study.
The percentile ranges of $\log$-likelihood in simulation studies are shown in Table~\ref{tab:simulikeliqrange}. When we have low event rates, VSI stands out from the baseline methods.
\begin{table}[ht]
\caption{Quantile range [0.10, 0.90] of $\log$-likelihood in Simulation Study}
% \begin{tabular}{llllll}
\begin{tabularx}{\columnwidth}{c *{5}{Y}}
\hline
Models      & Observed &       &       & Censored &       \\
            & 100\%    & 50\%  & 30\%  & 50\%     & 30\%  \\ \hline
AFT-Weibull & 1.55    & 1.67 & 1.87 & 0.26    & 0.28 \\
MLP         & 1.66    & 1.62 & 1.98 & 0.24    & 0.23 \\
VSI-NoQ     & 1.70    & 1.59 & 1.99 & 0.25    & 0.23 \\
VSI         & 1.67    & 1.60 & 1.68 & 0.23    & 0.27 \\ \hline
\end{tabularx}
\label{tab:simulikeliqrange}
\end{table}

{\bf Confidence Intervals for Raw C-Index} To better capture the differences in raw C-Index, the confidence intervals for each simulation strategy are calculated (See Table~\ref{tab:cisimu}). We have comparable performance with oracle methods with regarding to this metric. More illustrative metrics have been discussed in the text.
\begin{table}[ht]
\centering
\caption{Raw C-Index with confidence intervals (in parentheses) for simulation studies. RSF and DeepSurv do not provide intrinsic methods to calculate confidence intervals}
\label{tab:cisimu}
\begin{tabular}{@{}llll@{}}
\toprule
Event Rate  & 100\%                                                          & 50\%                                                           & 30\%                                                           \\ \midrule
CoxPH       & \begin{tabular}[c]{@{}l@{}}0.773\\ (0.768, 0.777)\end{tabular} & \begin{tabular}[c]{@{}l@{}}0.781\\ (0.775, 0.788)\end{tabular} & \begin{tabular}[c]{@{}l@{}}0.793\\ (0.785, 0.802)\end{tabular} \\
Coxnet      & \begin{tabular}[c]{@{}l@{}}0.776\\ (0.762, 0.789)\end{tabular} & \begin{tabular}[c]{@{}l@{}}0.784\\ (0.763, 0.805)\end{tabular} & \begin{tabular}[c]{@{}l@{}}0.760\\ (0.730, 0.791)\end{tabular} \\
AFT-Weibull & \begin{tabular}[c]{@{}l@{}}0.773\\ (0.768, 0.777)\end{tabular} & \begin{tabular}[c]{@{}l@{}}0.781\\ (0.775, 0.788)\end{tabular} & \begin{tabular}[c]{@{}l@{}}0.793\\ (0.785, 0.802)\end{tabular} \\
RSF         & 0.701                                                          & 0.718                                                          & 0.712                                                          \\
DeepSurv    & 0.772                                                          & 0.781                                                          & 0.793                                                          \\
MLP         & \begin{tabular}[c]{@{}l@{}}0.772\\ (0.767, 0.777)\end{tabular} & \begin{tabular}[c]{@{}l@{}}0.781\\ (0.775, 0.788)\end{tabular} & \begin{tabular}[c]{@{}l@{}}0.793\\ (0.784, 0.801)\end{tabular} \\
VSI-NoQ     & \begin{tabular}[c]{@{}l@{}}0.772\\ (0.768, 0.777)\end{tabular} & \begin{tabular}[c]{@{}l@{}}0.781\\ (0.774, 0.788)\end{tabular} & \begin{tabular}[c]{@{}l@{}}0.793\\ (0.784, 0.801)\end{tabular} \\
VSI         & \begin{tabular}[c]{@{}l@{}}0.773\\ (0.768, 0.777)\end{tabular} & \begin{tabular}[c]{@{}l@{}}0.781\\ (0.775, 0.788)\end{tabular} & \begin{tabular}[c]{@{}l@{}}0.793\\ (0.784, 0.801)\end{tabular} \\ \bottomrule
\end{tabular}
\end{table}

\begin{figure}[ht]
\centering
\begin{tabular}{c@{}c}
\subfloat[]{\includegraphics[width=0.49\linewidth]{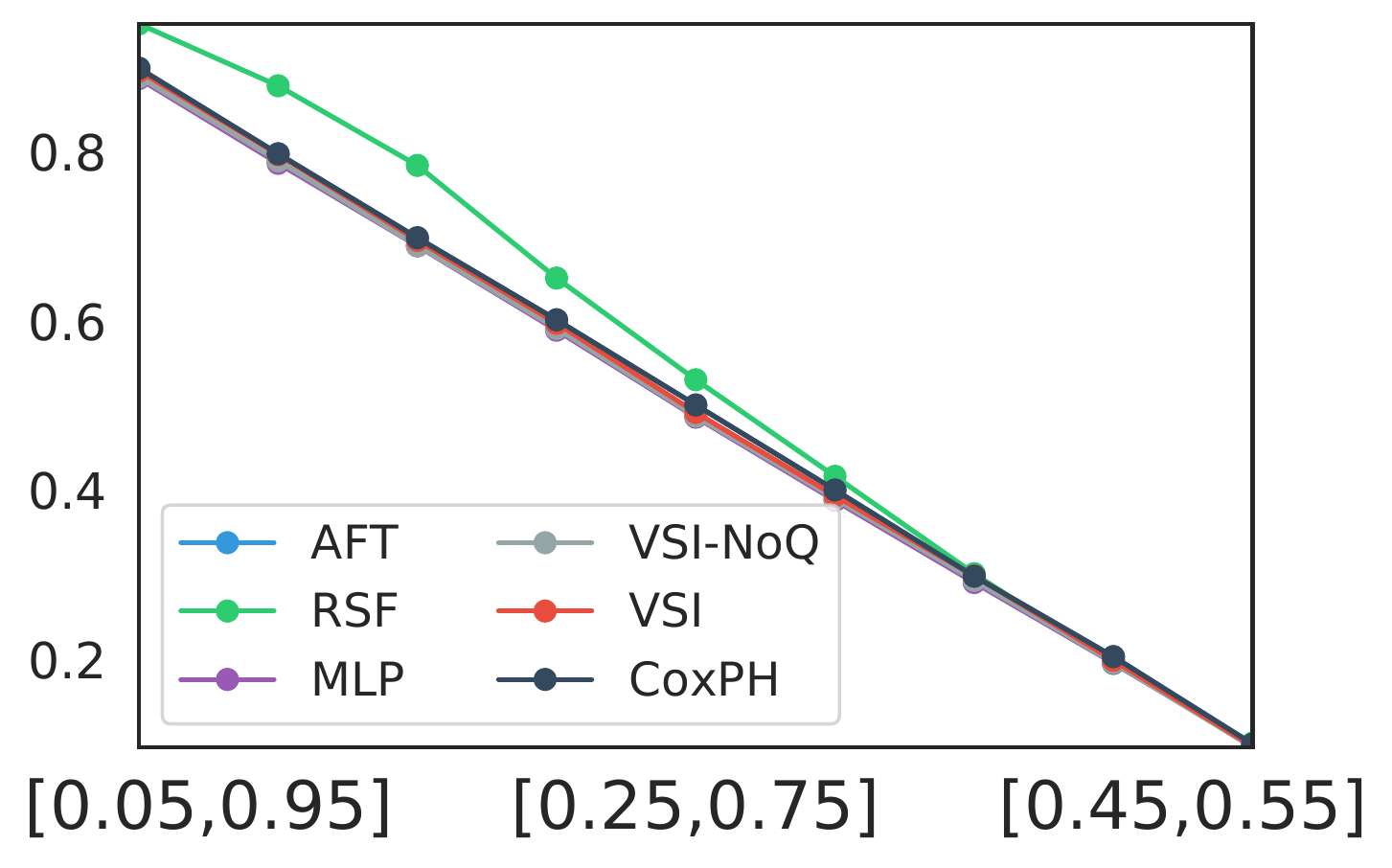}} &\\
\subfloat[]{\includegraphics[width=0.49\linewidth]{plots/simu_er50_e_coverage.pdf}} & \subfloat[]{\includegraphics[width=0.49\linewidth]{plots/simu_er50_c_coverage.pdf}}\\
\subfloat[]{\includegraphics[width=0.49\linewidth]{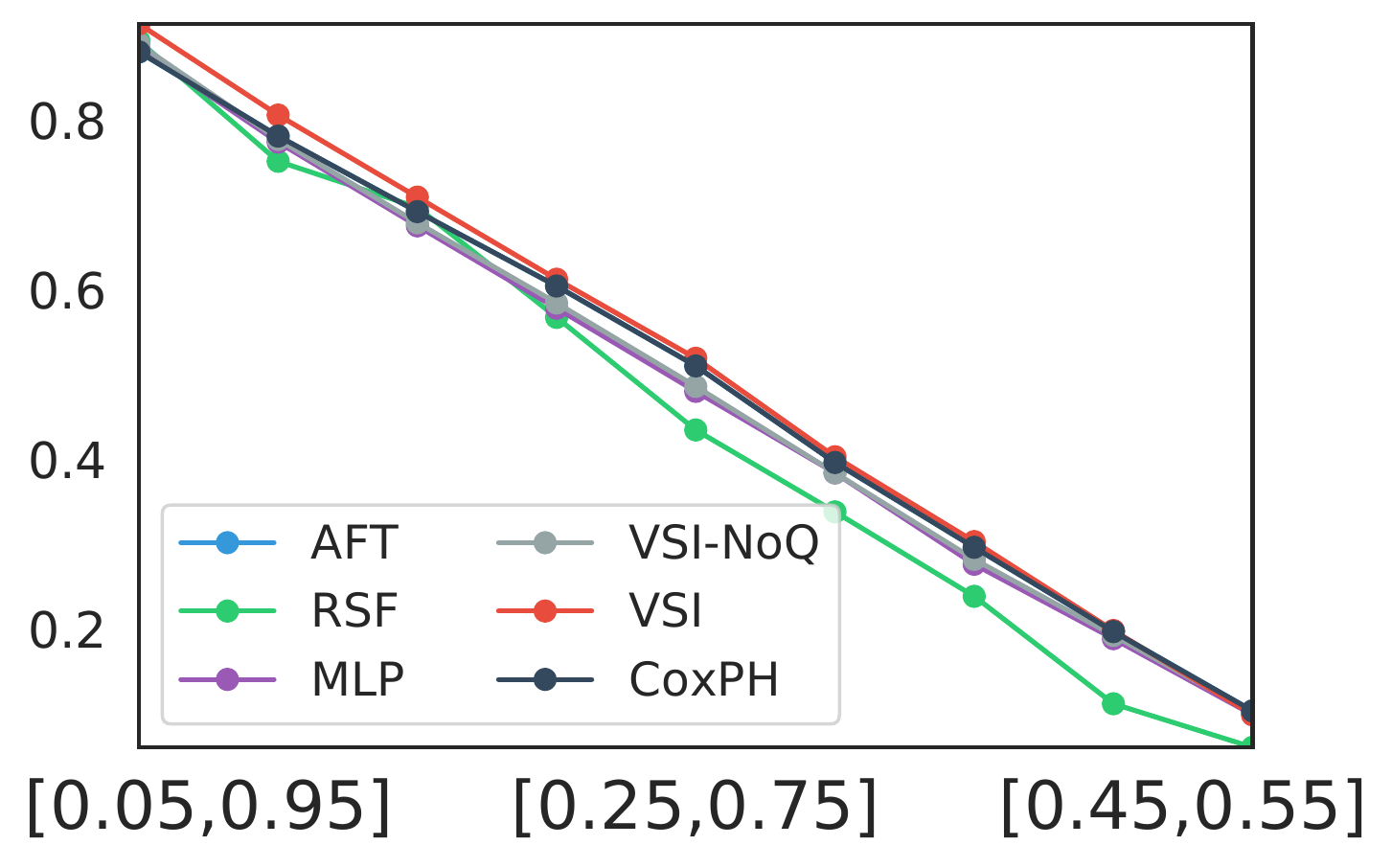}}&
\subfloat[]{\includegraphics[width=0.49\linewidth]{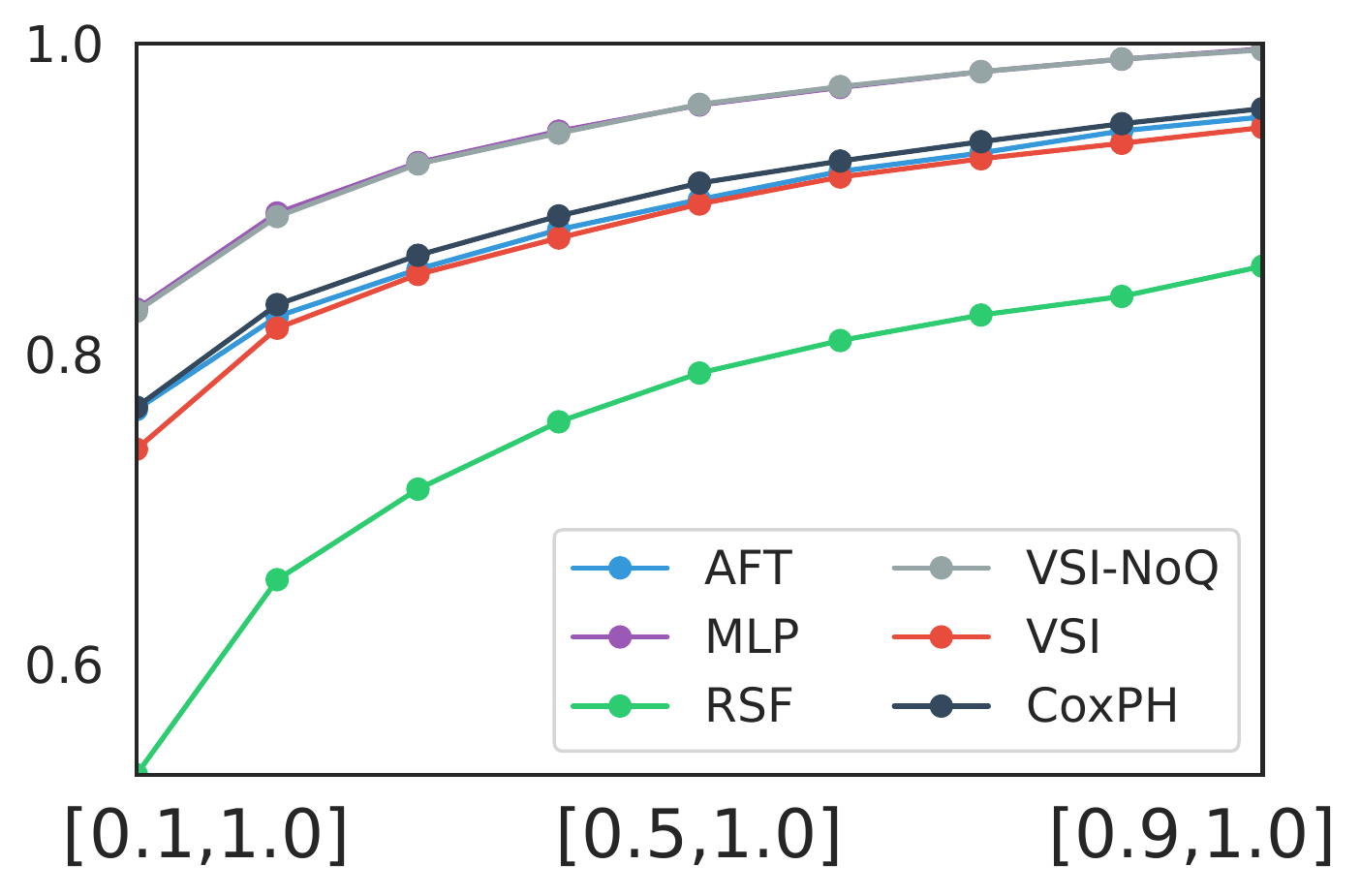}}\\
\end{tabular}
\caption{Testing datasets Coverage rate for simulation datasets. $100\%$ event rate (a), $50\%$ event rate (observed:b, censored:c), $30\%$ event rate (observed:d, censored:e)}
\label{fig:simucover}
\end{figure}

{\bf Coverage Rate}
To visualize the proportion of observed time is covered in the predicted personalized time-to-event distributions, we calculated the coverage rate for each percentile ranges. Our model balanced between events and censoring coverage the best among all methods. We compared the coverage with CoxPH, AFT-Weibull, VSI-NoQ, MLP and RSF. CoxPH could serves as the reference, since the simulation study is based on CoxPH assumptions entirely. The results are shown in Figure~\ref{fig:simucover}.
%

% \clearpage
\section{Additional Results for Real Datasets}

{\bf Comparison of Weighted Average and Median}
In Table~\ref{tab:realcicompare}, we compared the C-Index calculated with weighted average and median for real datasets, with weighted average higher in general. In simulation studies, those two numbers are similar to each other.
\begin{table}[h!]
\centering
\caption{Comparison median and weighted average for predicted time-to-event distribution with C-index in real datasets}
\begin{tabular}{@{}llll@{}}
\toprule
C-Index          & \textsc{FLCHAIN} & \textsc{SUPPORT} & \textsc{SEER}  \\ \midrule
Median           & 0.774   & \textbf{0.833}   & 0.814 \\
Weighted Average & \textbf{0.792}   & 0.823   & \textbf{0.826} \\ \bottomrule
\end{tabular}
\label{tab:realcicompare}
\end{table}

\begin{figure}[ht!]
\centering
\begin{tabular}{c@{}c}
\subfloat[]{\includegraphics[width=0.49\linewidth]{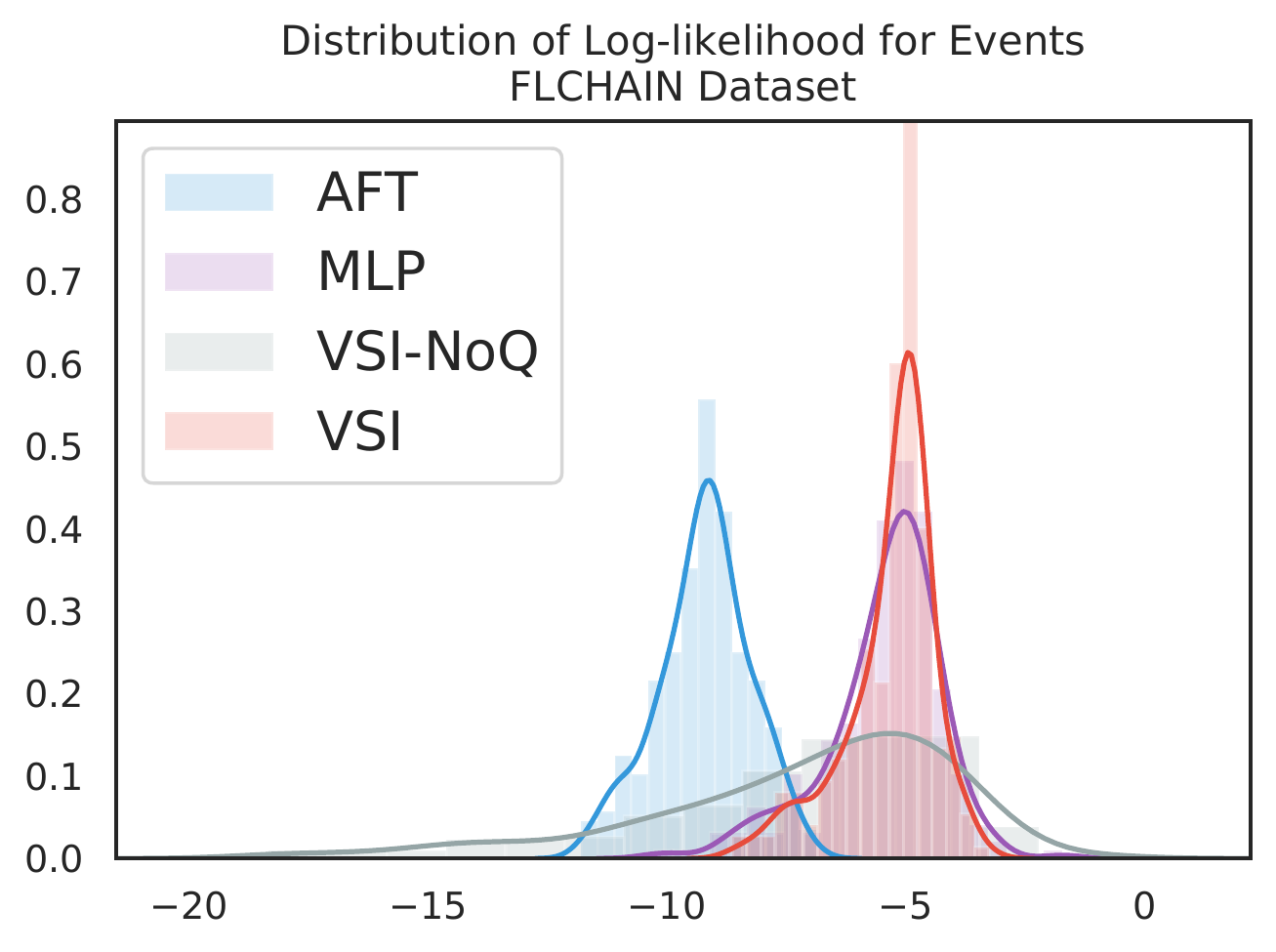}} &
\subfloat[]{\includegraphics[width=0.49\linewidth]{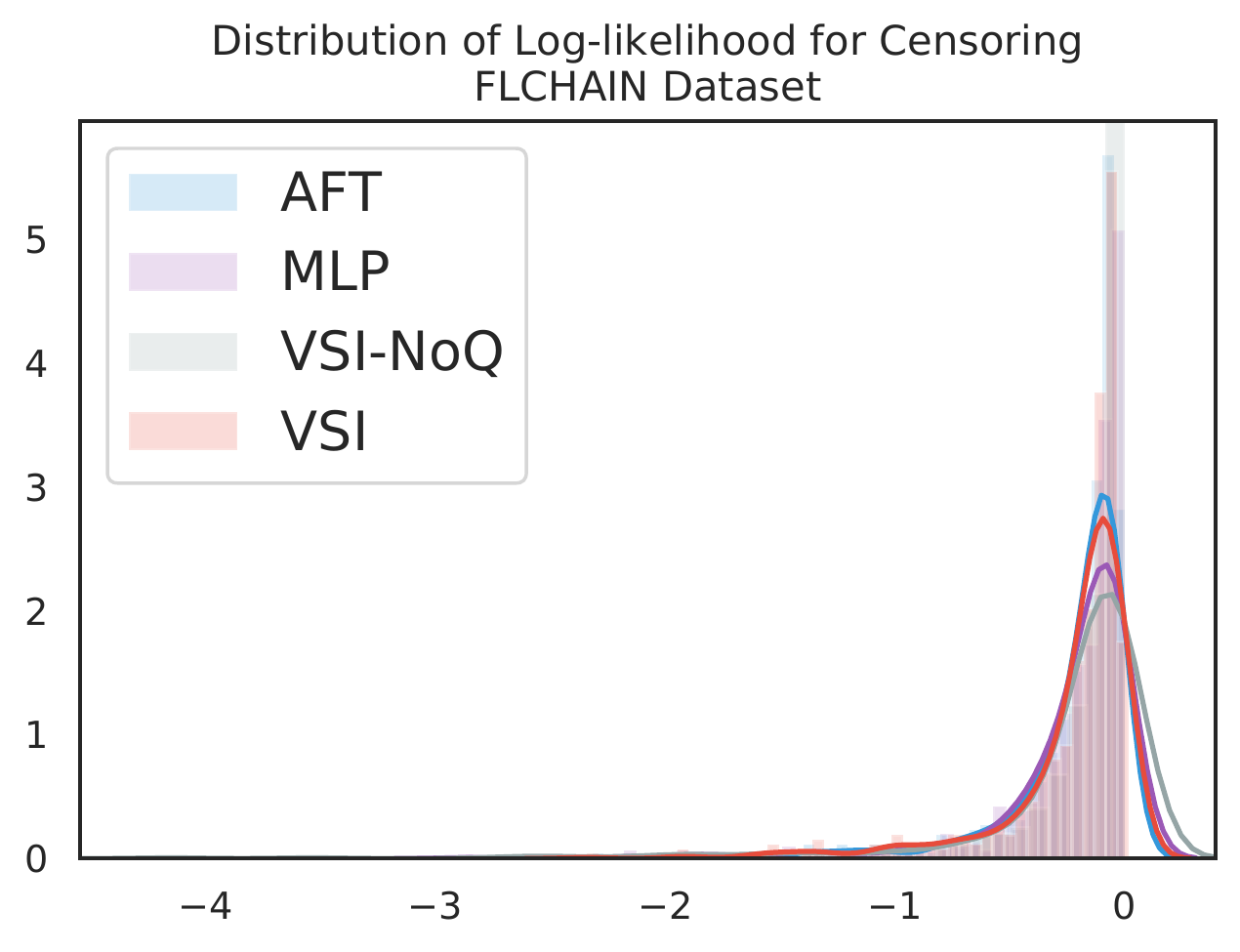}}\\ \subfloat[]{\includegraphics[width=0.49\linewidth]{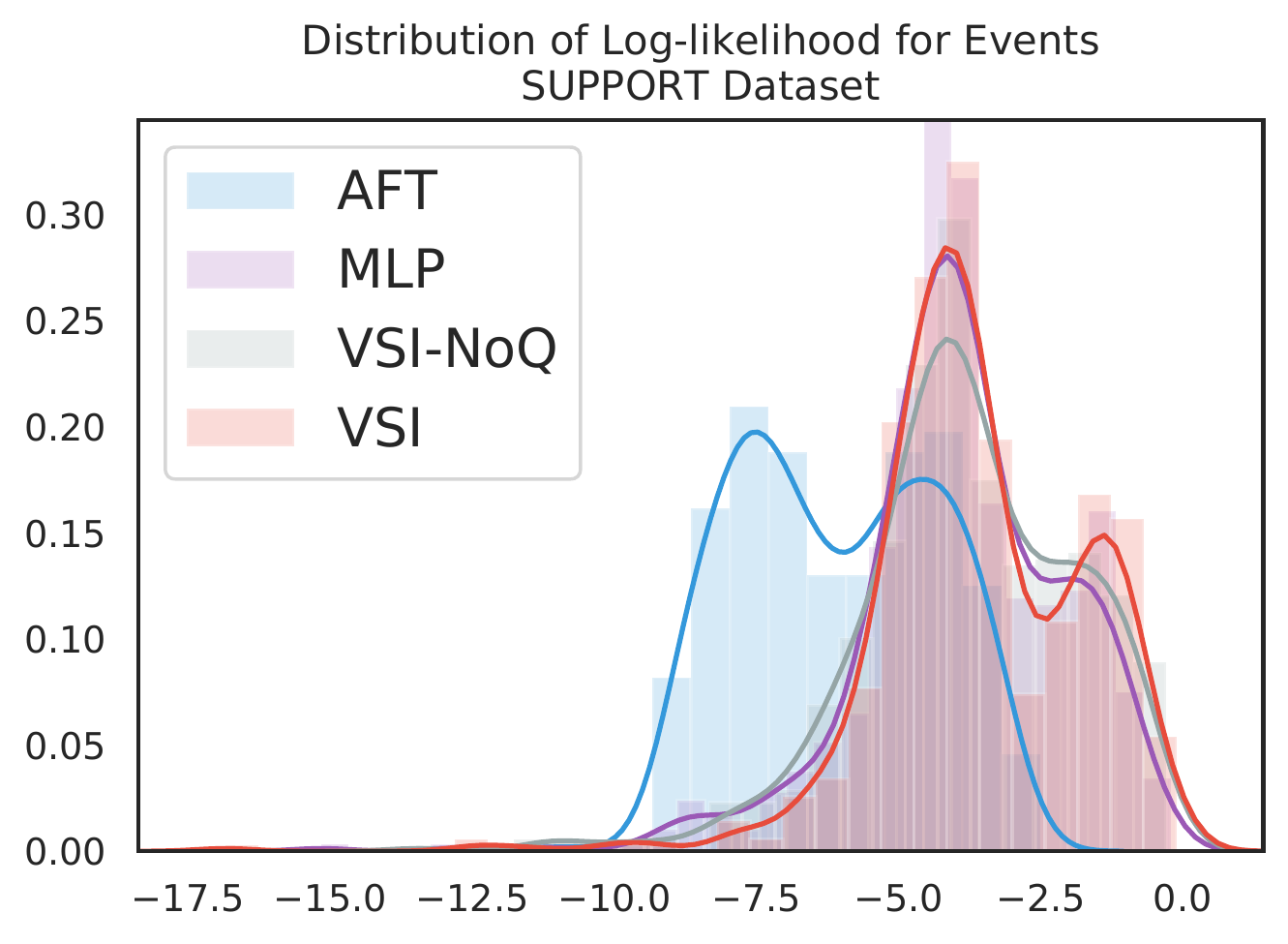}}&
\subfloat[]{\includegraphics[width=0.49\linewidth]{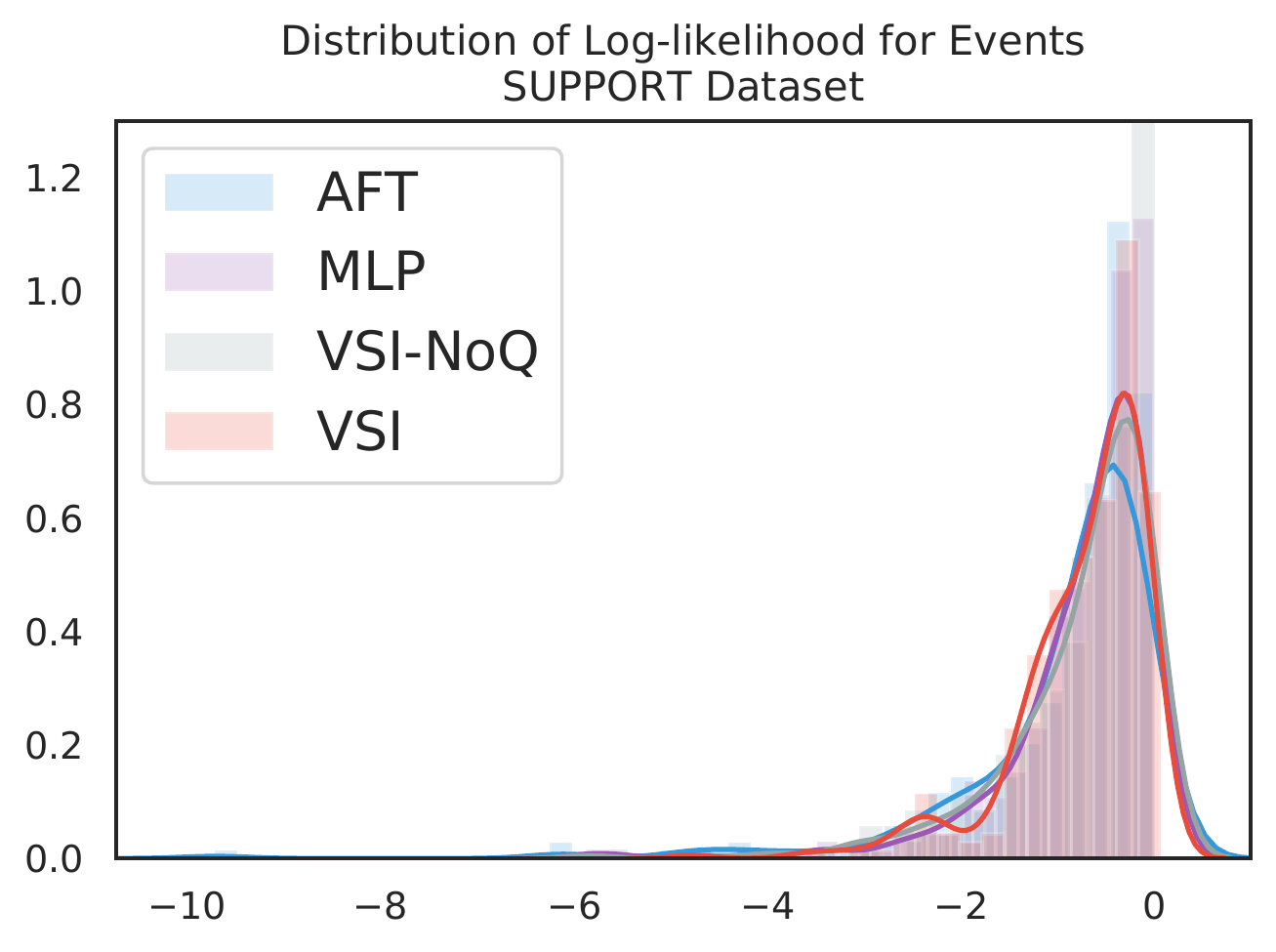}}\\
\subfloat[]{\includegraphics[width=0.49\linewidth]{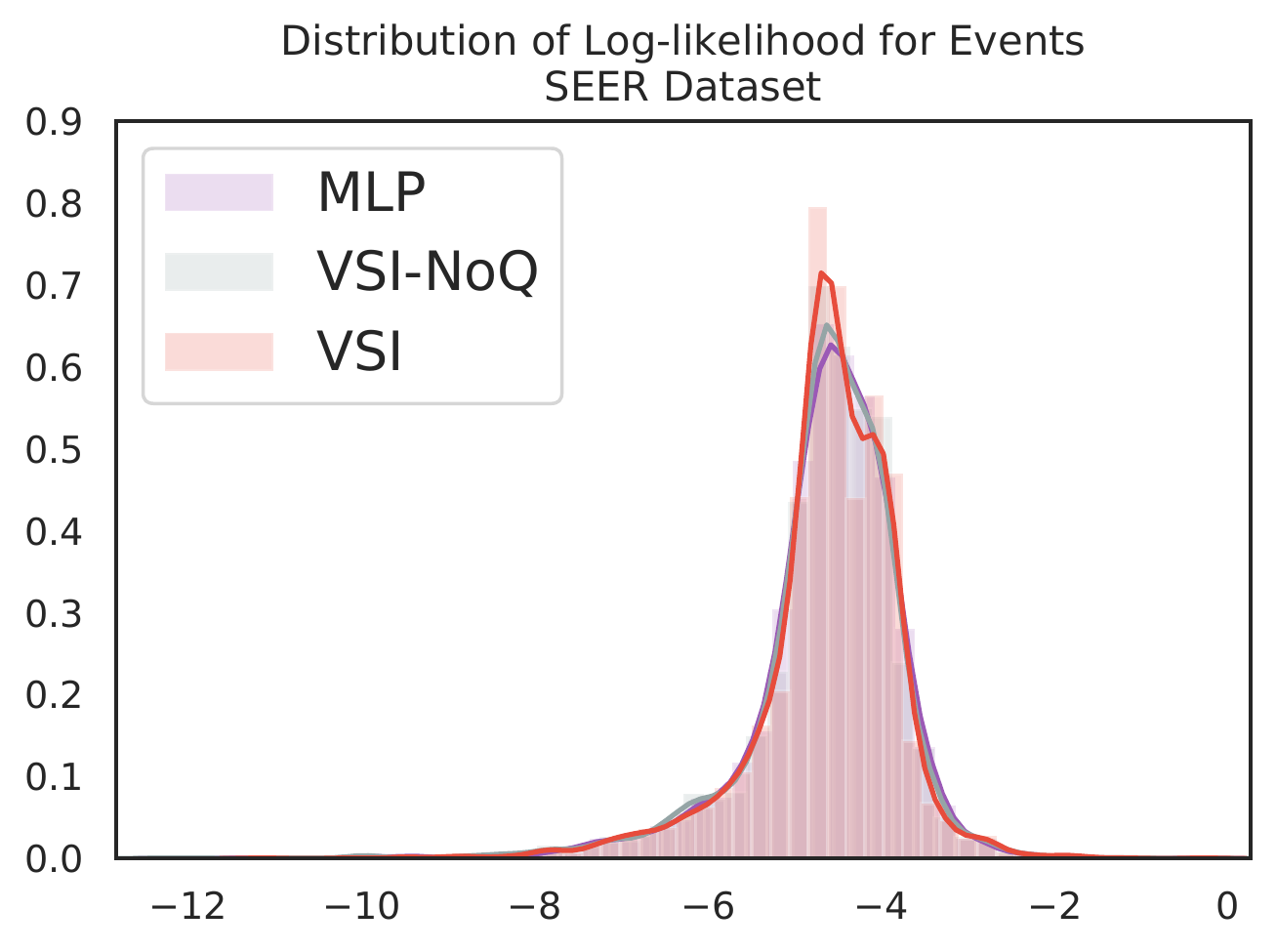}}&
\subfloat[]{\includegraphics[width=0.49\linewidth]{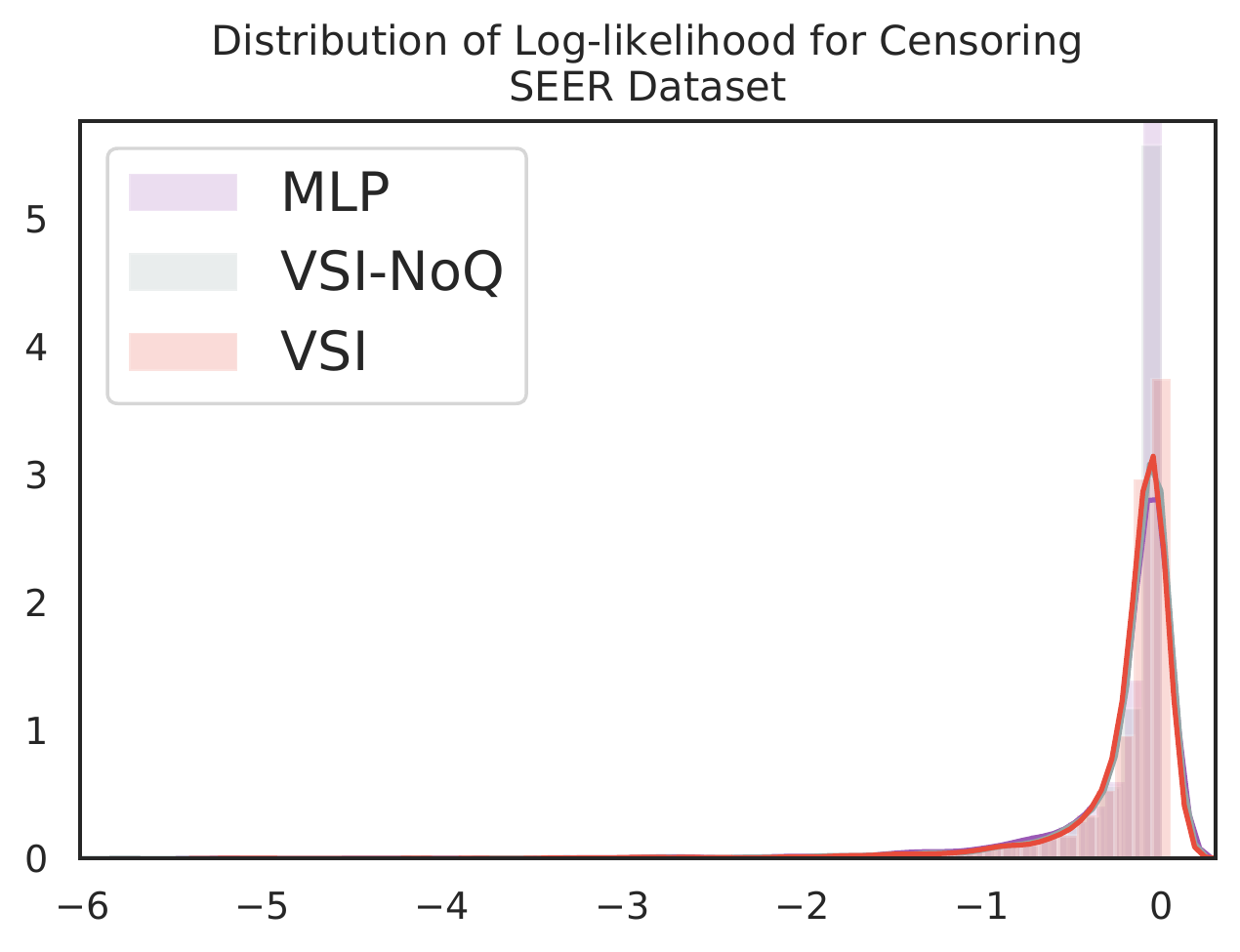}}\\
\end{tabular}
\caption{Likelihood distributions for real datasets. \textsc{FLCHAIN} (observed: a, censored: b), \textsc{SUPPORT}  (observed: c, censored: d), \textsc{SEER} (observed: e, censored: f)}
\label{fig:reallikeli-all}
\end{figure}

\vspace{3pt}
{\bf Distribution for Log-likelihood} Log-likelihood distribution for real datasets shown in Figure~\ref{fig:reallikeli-all}. Similar to the case discussed in the text, VSI performs consistently better in all three realworld datasets.

\vspace{3pt}
{\bf Confidence Intervals for Raw C-Index} As shown in Table~\ref{tab:cireal}, raw C-Index for VSI is significantly better than other method with relatively tight confidence intervals.
\begin{table}[ht]
\centering
\caption{Raw C-Index with confidence intervals (in parentheses) for Real Data. NA indicates the corresponding evaluation metric cannot be evaluated. RSF and DeepSurv do not provide intrinsic methods to calculate confidence intervals.}
\label{tab:cireal}
\vspace{1mm}
\begin{tabular}{@{}llll@{}}
\toprule
Event Rate  & FLCHAIN                                                        & SUPPORT                                                        & SEER                                                            \\ \midrule
Coxnet      & \begin{tabular}[c]{@{}l@{}}0.790\\ (0.761, 0.820)\end{tabular} & \begin{tabular}[c]{@{}l@{}}0.784\\ (0.763, 0.805)\end{tabular} & \begin{tabular}[c]{@{}l@{}}0.819\\ (0.812, 0.826)\end{tabular}  \\
AFT-Weibull & \begin{tabular}[c]{@{}l@{}}0.792\\ (0.763, 0.821)\end{tabular} & \begin{tabular}[c]{@{}l@{}}0.797\\ (0.782, 0.813)\end{tabular} & NA                                                              \\
RSF         & 0.771                                                          & 0.751                                                          & 0.796                                                           \\
DeepSurv    & 0.785                                                          & 0.678                                                          & NA                                                              \\
MLP         & \begin{tabular}[c]{@{}l@{}}0.751\\ (0.722, 0.781)\end{tabular} & \begin{tabular}[c]{@{}l@{}}0.811\\ (0.791, 0.831)\end{tabular} & \begin{tabular}[c]{@{}l@{}}0.811\\ (0.803, 0.818)\end{tabular}  \\
VSI-NoQ     & \begin{tabular}[c]{@{}l@{}}0.745\\ (0.714, 0.777)\end{tabular} & \begin{tabular}[c]{@{}l@{}}0.824\\ (0.804, 0.843)\end{tabular} & \begin{tabular}[c]{@{}l@{}}0.809\\ (0.802,  0.817)\end{tabular} \\
VSI         & \begin{tabular}[c]{@{}l@{}}{\bf 0.792}\\ (0.762, 0.821)\end{tabular} & \begin{tabular}[c]{@{}l@{}}{\bf 0.827}\\ (0.809, 0.846)\end{tabular} & \begin{tabular}[c]{@{}l@{}}{\bf 0.826}\\ (0.819, 0.833)\end{tabular}  \\ \bottomrule
\end{tabular}
\end{table}

\begin{figure}[ht]
\centering
\begin{tabular}{c@{}c}
\subfloat[]{\includegraphics[width=0.49\linewidth]{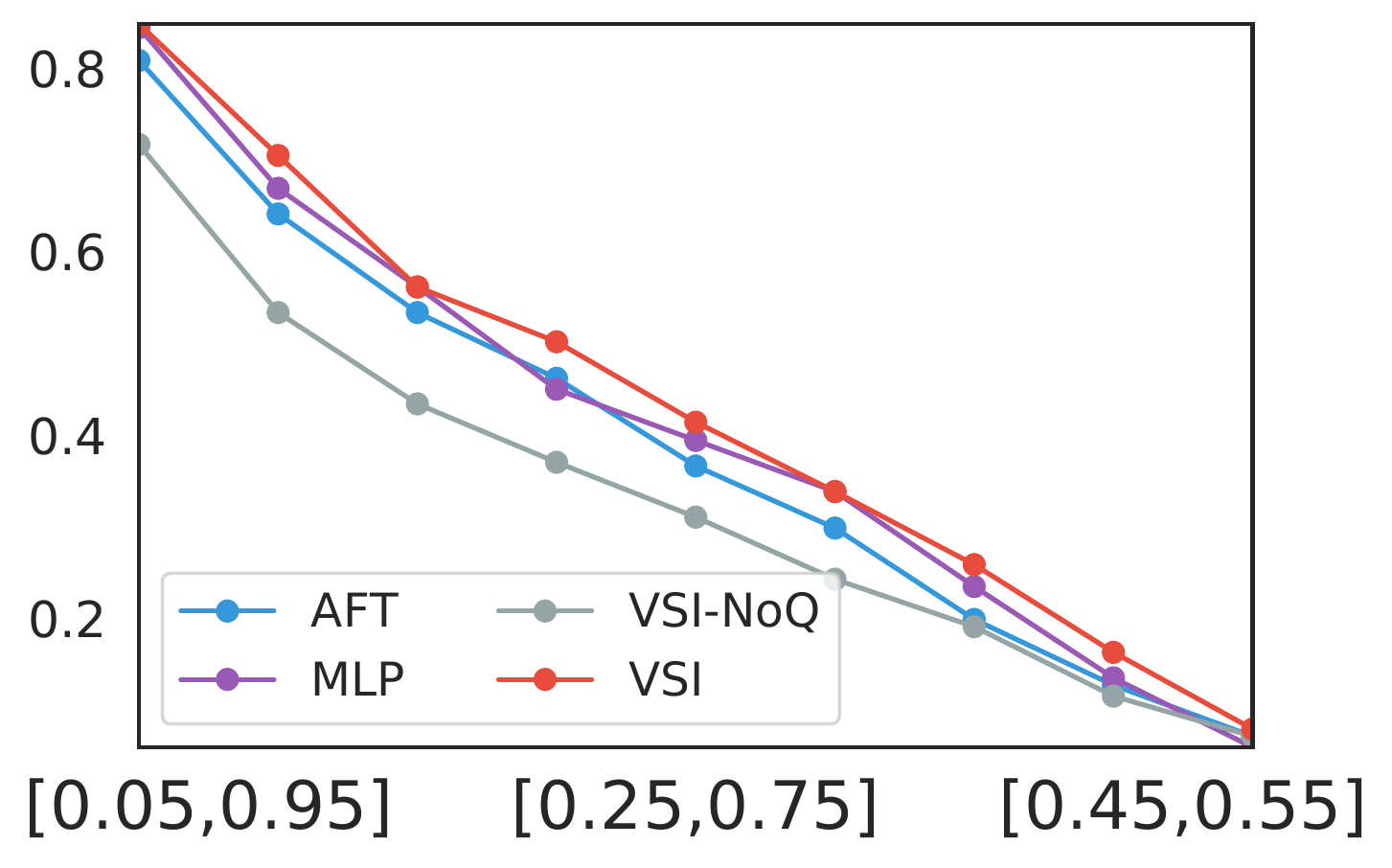}} &
\subfloat[]{\includegraphics[width=0.49\linewidth]{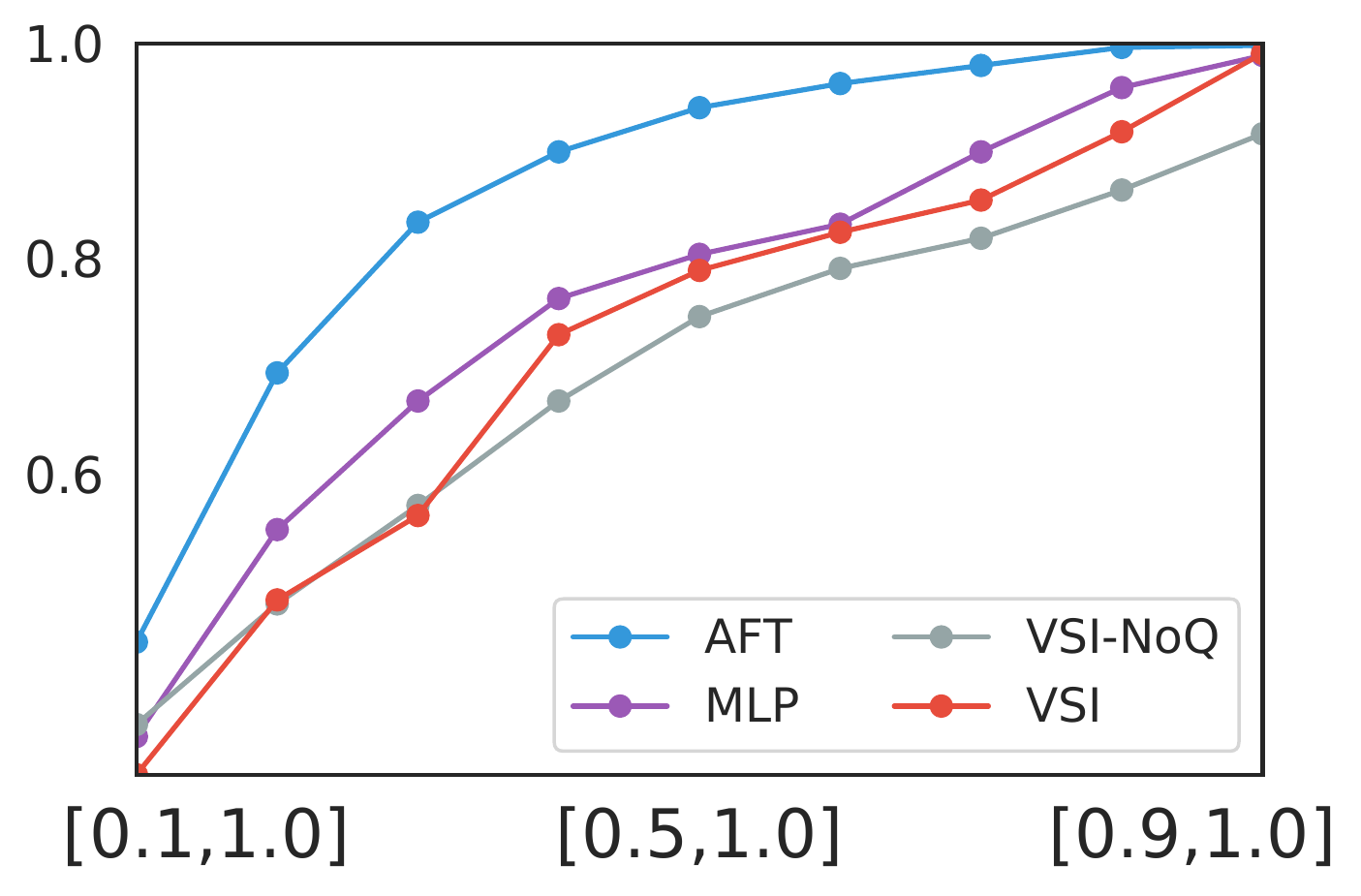}}\\ \subfloat[]{\includegraphics[width=0.49\linewidth]{plots/support_e_coverage.pdf}}&
\subfloat[]{\includegraphics[width=0.49\linewidth]{plots/support_c_coverage.pdf}}\\
\subfloat[]{\includegraphics[width=0.49\linewidth]{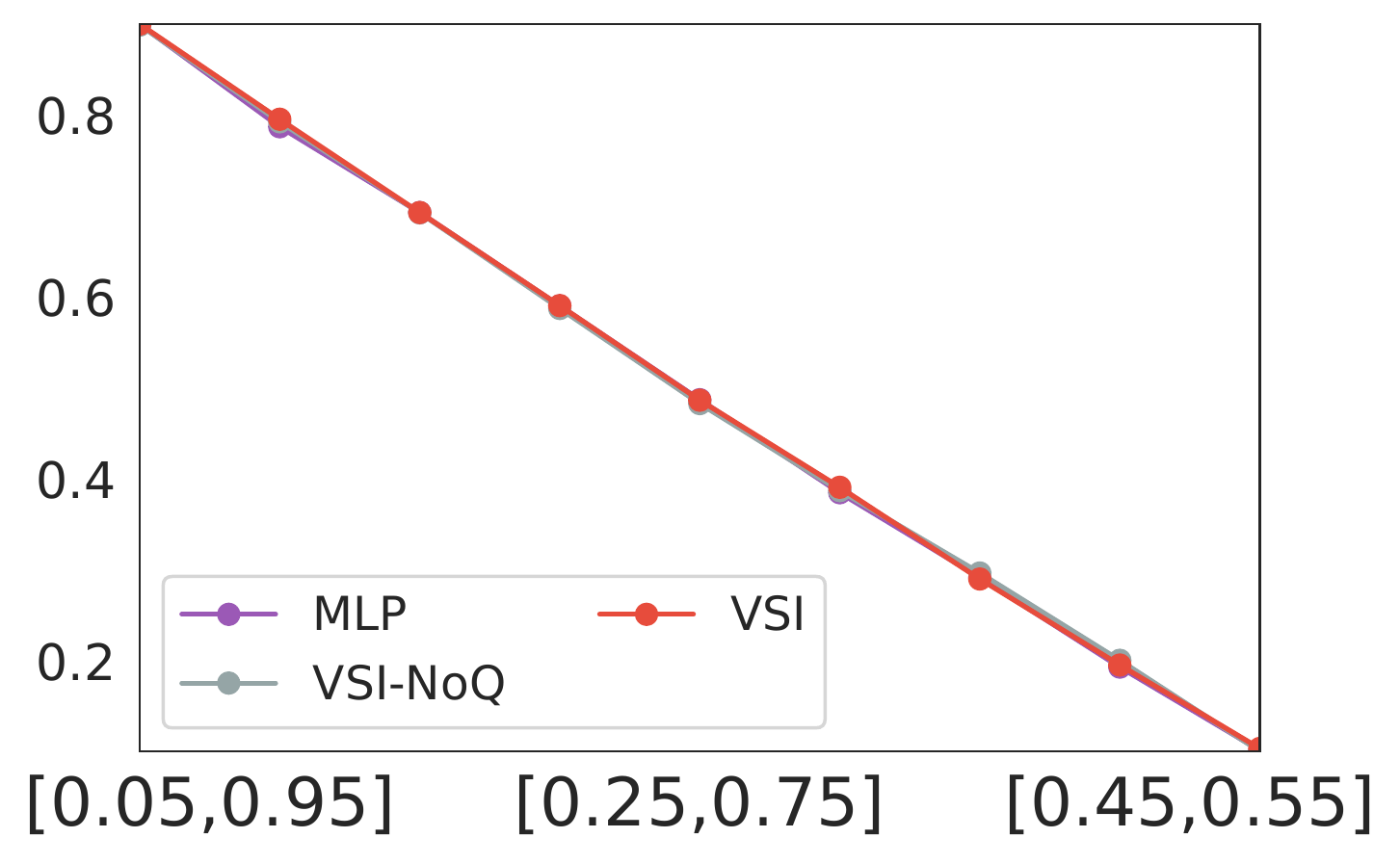}}&
\subfloat[]{\includegraphics[width=0.49\linewidth]{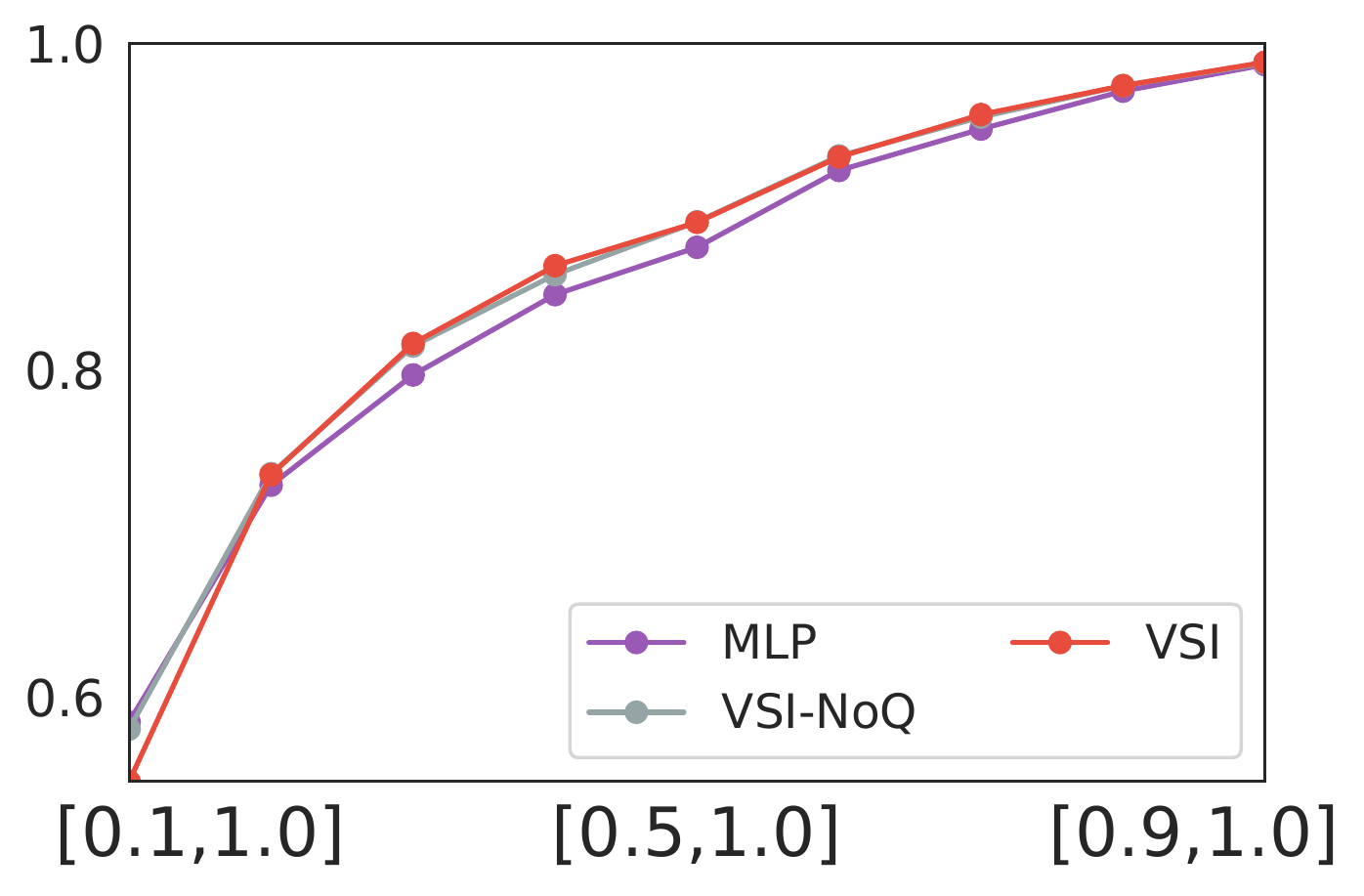}}\\
\end{tabular}
\caption{Coverage rates for real datasets. \textsc{FLCHAIN} (observed: a, censored: b), \textsc{SUPPORT}  (observed: c, censored: d), \textsc{SEER} (observed: e, censored: f)}
\label{fig:realcover}
\end{figure}

\vspace{3pt}
{\bf Coverage rate for real datasets} Cover rate for real datasets shown in Figure~\ref{fig:realcover}. VSI has relative high coverage for both events and censoring in all three datasets, which similar to the case we have in simulation datasets.

For the comparison models, VSI, MLP and VSI-NoQ and AFT-Weibull are distribution based and could give all statistics in the simulation studies. For RSF and CoxPH, which could calculate the estimated cumulative hazards, the corresponding survival function and time-to-event could be calculated accordingly. For DeepSurv and Coxnet, only raw C-Index can be given.

\end{document}